\documentclass[lettersize,journal]{IEEEtran}
\usepackage{amsmath,amsfonts}
\usepackage{algorithmic}
\usepackage{algorithm}
\usepackage{array}
\usepackage[caption=false,font=footnotesize]{subfig}
\usepackage{textcomp}
\usepackage{stfloats}
\usepackage{url}
\usepackage{verbatim}
\usepackage{graphicx}
\usepackage{cite}
\usepackage{multirow}
\usepackage{booktabs}

\usepackage{hyperref}

\usepackage{xcolor}
\definecolor{BLUE}{rgb}{0,0,1} 
\usepackage{colortbl}

\usepackage{amssymb}

\usepackage{subcaption}

\usepackage{makecell}

\begin{document}


\title{MTOR: Generalizable AI-Generated Video Detection with Multimodal Semantics and Temporal Over-Regularity}

\author{Hang~Wang,~\IEEEmembership{Member,~IEEE,}
        Chao~Shen,~\IEEEmembership{Fellow,~IEEE,}
        Lei~Zhang,~\IEEEmembership{Fellow,~IEEE,} 
        and~Zhi-Qi~Cheng,~\IEEEmembership{Senior Member,~IEEE}
\thanks{H. Wang is with Xi’an Jiaotong University, Xi’an, China, and The Hong Kong Polytechnic University, Hong Kong, China (e-mail: cshangwang@xjtu.edu.cn).}
\thanks{C. Shen is with Xi’an Jiaotong University, Xi'an, China (e-mail: chaoshen@mail.xjtu.edu.cn).}
\thanks{L. Zhang is with The Hong Kong Polytechnic University, Hong Kong, China (e-mail: cslzhang@comp.polyu.edu.hk).}
\thanks{Z.-Q. Cheng is with the Language Technologies Institute, Carnegie Mellon University, Pittsburgh, PA, USA (e-mail: zhiqic@cs.cmu.edu).}
\thanks{Corresponding authors: Chao Shen; Zhi-Qi Cheng.}
}

\markboth{Journal of \LaTeX\ Class Files,~Vol.~14, No.~8, August~2021}%
{Shell \MakeLowercase{\textit{et al.}}: A Sample Article Using IEEEtran.cls for IEEE Journals}


\maketitle

\begingroup
\renewcommand\thefootnote{}
\footnotetext{\footnotesize
This work has been submitted to the IEEE for possible publication.
Copyright may be transferred without notice, after which this version may no longer be accessible.}
\addtocounter{footnote}{-1}
\endgroup

\begin{abstract}
    The rapid evolution of video generation has narrowed the perceptual gap between authentic and synthetic videos, making generalizable AI-generated video detection increasingly challenging.
    Existing detectors predominantly rely on visual representations, leaving caption-derived textual semantics underexplored.
    Meanwhile, temporal regularity in fine-grained visual representations has received limited attention.
    We find that caption-derived textual representations provide complementary discriminative cues to global visual representations.
    Our analysis further reveals that AI-generated videos exhibit stronger temporal persistence and lower temporal variability, a pattern we term temporal over-regularity (TOR).
    Based on these findings, we propose MTOR with a multimodal branch and a TOR component. 
    The multimodal branch integrates global visual and caption-derived textual representations, while the TOR component models temporal over-regularity at three levels: coarse inter-frame continuity, fine-grained token correspondence, and frame-to-video stability.
    Extensive evaluations on five benchmarks covering 46 generator variants demonstrate state-of-the-art overall performance against 16 representative baselines, while robustness experiments confirm strong resilience to twelve real-world video perturbations.
    Code and models will be released.\footnote{\url{https://github.com/hwang-cs-ime/MTOR}}
\end{abstract}


\section{Introduction}
The rapid evolution of generative artificial intelligence has spurred unprecedented breakthroughs in video synthesis.
Driven by scaling laws and advanced diffusion-based architectures, state-of-the-art video generation frameworks such as Sora~\cite{brooks2024video}, Runway Gen-4~\cite{runwayGen42025} and Kling~\cite{klingai2025} are capable of producing high-fidelity videos with photorealistic visual quality, fluent motion, and coherent spatiotemporal scenes.
These advances have made AI-generated videos increasingly difficult to visually distinguish from authentic videos.
Meanwhile, the widespread availability of these sophisticated generation techniques brings severe societal risks.
Maliciously fabricated videos can easily propagate misinformation~\cite{bao2018towards, guo2024pulid, zheng2023out, ijcai2021p157, zhao2023diffswap, oorloff2024avff, li2023learning}, undermine public trust~\cite{barrett2023identifying, sharma2023survey} in digital media, and even pose potential ethical and security threats. 
Consequently, developing a generalizable detection paradigm to identify AI-generated videos has become increasingly important for advancing multimedia forensics and safeguarding societal trust.

To mitigate these threats, recent studies have explored diverse temporal and spatiotemporal modeling strategies for AI-generated video detection.
DeCoF~\cite{Ma2024DetectingAV} focuses on inter-frame consistency to capture temporal artifacts that generalize across generators, while ReStraV~\cite{interno2025aigenerated} analyzes the geometry of visual representation trajectories through temporal curvature and distance.
D3~\cite{Zheng_2025_ICCV} further characterizes second-order temporal variation, whereas NSG-VD~\cite{zhang2025NSGVD} models spatiotemporal dynamics through conservation constraints on probability flow.
STALL~\cite{hayun2026stall} takes a probabilistic perspective by jointly modeling spatial likelihoods of individual frames and temporal likelihoods of inter-frame transitions.
Despite these advances, most existing approaches base their predictions primarily on visual representations, leaving caption-derived textual semantics largely unexplored.
Such reliance on visual representations may overlook complementary discriminative cues conveyed by caption-derived textual semantics.
Meanwhile, existing temporal and spatiotemporal methods characterize video dynamics from diverse perspectives, yet the temporal regularities encoded in fine-grained visual representations have not been fully exploited for generalizable detection.

To address these limitations, we investigate AI-generated video detection from two complementary perspectives: video-level multimodal semantics and fine-grained temporal regularity.
First, we examine the contribution of caption-derived textual representations when combined with global visual representations.
Our empirical results show that their combination consistently outperforms either representation alone, indicating that caption-derived semantics provide discriminative cues complementary to visual representations.
Second, we analyze fine-grained visual dynamics at three levels: coarse inter-frame continuity, fine-grained token correspondence, and frame-to-video stability.
Across these levels, AI-generated videos consistently exhibit stronger temporal persistence and reduced temporal variability than real-world videos.
We term this tendency \emph{temporal over-regularity} (TOR).

Based on these observations, we propose \textbf{MTOR}, a multimodal AI-generated video detector that jointly models video-level \textbf{M}ultimodal semantics and \textbf{T}emporal \textbf{O}ver-\textbf{R}egularity.
Given an input video, a video encoder extracts a global visual representation, while a video captioning model generates a textual description that is subsequently encoded into a global textual representation.
The multimodal branch combines the visual and textual representations to produce a multimodal detection score.
In parallel, the TOR component explicitly models the three levels of temporal regularity using six descriptors that capture the strength and variation of adjacent-frame relations, together with the stability of frame-to-video similarity.
These descriptors are subsequently mapped to a TOR detection score.
Finally, the multimodal and TOR scores are standardized using statistics estimated from the source-training set and fused to obtain the final MTOR score, without requiring any target-domain data.

In summary, this work makes three key contributions.
\begin{enumerate}
    \item We propose MTOR, a multimodal framework for generalizable AI-generated video detection. Its multimodal branch integrates global visual and caption-derived textual representations derived from the same input video, introducing complementary textual semantics without requiring external textual input.

    \item We reveal that AI-generated videos exhibit stronger temporal persistence and reduced temporal variability than real videos, a pattern we term temporal over-regularity (TOR). To model this pattern, we develop a TOR component with six temporal descriptors spanning coarse inter-frame continuity, fine-grained token correspondence, and frame-to-video stability.

    \item Extensive experiments on five benchmarks covering 46 video generator variants demonstrate strong cross-generator performance against 16 representative baselines. Ablation and robustness studies further verify the complementarity of the multimodal branch and TOR component, as well as the robustness of MTOR under twelve real-world video perturbations.
\end{enumerate}

\section{Related Work}
\subsection{Deepfake Video Detection}

Deepfake video detection has gradually shifted from frame-level artifact analysis toward exploiting temporal and spatiotemporal inconsistencies across video frames. 
Early video-based methods therefore focused on exposing inter-frame inconsistencies. 
STIL~\cite{10.1145/3474085.3475508} integrates spatial inconsistencies with adjacent-frame temporal differences, while FTCN~\cite{zheng2021exploring} suppresses spatial modeling to emphasize temporal features and captures long-range temporal coherence with a Temporal Transformer. 
Subsequent studies further model temporal dependencies over multiple ranges. 
MRE-Net~\cite{pang2023mre} employs multi-rate sampling to capture short- and long-term spatiotemporal inconsistencies, whereas ISTVT~\cite{zhao2023istvt} decomposes self-attention into spatial and temporal dimensions and emphasizes inter-frame temporal inconsistencies.
TALL~\cite{xu2023tall} rearranges consecutive frames into a compact thumbnail for efficient spatiotemporal modeling, while MINTIME~\cite{10547206} introduces identity-aware modeling for videos containing multiple identities and varying face sizes.

As deepfake generation techniques become increasingly realistic, improving generalization to unseen manipulations has received growing attention. 
AltFreezing~\cite{wang2023altfreezing} alternately freezes spatial- and temporal-related weights to prevent excessive reliance on either artifact type, while DIP~\cite{nie2024dip} models directional spatiotemporal inconsistencies and introduces generalization-oriented training strategies. 
Beyond conventional RGB-level temporal cues, StyleFlow~\cite{choi2024exploiting} analyzes temporal variations of facial style latent vectors, and recent work~\cite{Kim_2025_ICCV} extends temporal analysis to the frequency domain. 
Yan \emph{et al.}~\cite{yan2025generalizing} identify Facial Feature Drift as a transferable temporal forgery artifact and construct hard negative samples through video-level blending, whereas FakeSTormer~\cite{nguyen2025vulnerability} introduces auxiliary spatial and temporal vulnerability branches to improve generalization and interpretability. 
Collectively, these studies seek more transferable temporal representations and learning strategies for generalized deepfake detection.

Multimodal cues have also been investigated for deepfake video detection.
Feng et al.~\cite{feng2023self} formulate video forensics as self-supervised anomaly detection based on audio-visual synchronization, while AVFF~\cite{oorloff2024avff} learns audio-visual correspondences through self-supervised contrastive and reconstruction objectives before supervised fine-tuning. 
These studies demonstrate that inconsistencies in audio-visual synchronization and correspondence can help distinguish manipulated from authentic videos.
Nevertheless, most deepfake video detectors are designed for facial manipulation and primarily exploit face-specific spatial, temporal, or audio-visual cues. 
Their problem setting therefore differs from open-domain AI-generated video detection, where entire videos may be synthesized by diverse generation models and span a wide range of scenes, objects, actions, and motion patterns.

\subsection{AI-Generated Video Detection}
Unlike deepfake detection, which primarily targets facial manipulation, AI-generated video detection addresses fully synthesized videos beyond the facial domain, making generalization to unseen video generators particularly challenging.
Early studies approached this problem by modeling spatial and temporal artifacts in generated videos.
AIGVDet~\cite{bai2024ai} employs separate RGB and optical-flow branches to capture spatial abnormalities and temporal inconsistencies, respectively, and combines their frame-level predictions through decision-level fusion for video-level detection.
DeMamba~\cite{chen2024demamba} further models local spatiotemporal inconsistencies across different regions using a structured state-space model.
However, spatial artifacts are often generator-specific.
Motivated by this observation, DeCoF~\cite{Ma2024DetectingAV} reduces the influence of spatial artifacts during feature learning and focuses on inter-frame consistency to improve generalization to unseen generators.
Together, these studies highlight spatial and temporal artifacts as important forensic cues for AI-generated video detection.

More recent studies have examined video dynamics from increasingly diverse perspectives.
ReStraV~\cite{interno2025aigenerated} analyzes frame trajectories in a pretrained visual representation space and characterizes their geometry through temporal curvature and stepwise distance.
D3~\cite{Zheng_2025_ICCV} derives second-order temporal features from first-order inter-frame differences and performs training-free detection using statistics of second-order central differences.
In a different direction, NSG-VD~\cite{zhang2025NSGVD} takes a physics-driven approach to modeling spatiotemporal video evolution through conservation constraints on probability flow.
It defines the Normalized Spatiotemporal Gradient (NSG) as the ratio of spatial probability gradients to temporal density changes, and detects AI-generated videos by comparing NSG distributions between test and real videos.
STALL~\cite{hayun2026stall} adopts a training-free probabilistic framework that jointly models the spatial likelihood of individual frames and the temporal likelihood of frame-to-frame transitions.

Overall, existing methods span a wide range of visual-domain formulations, from spatial and temporal artifacts to representation geometry, higher-order dynamics, physics-driven spatiotemporal modeling, and spatial-temporal likelihoods.
In contrast, MTOR extends beyond purely visual formulations by incorporating a caption-derived textual representation of the same video.
It further characterizes temporal over-regularity through coarse inter-frame continuity, fine-grained token correspondence, and frame-to-video stability.

\begin{table}[!ht]
    \centering
    \caption{Performance comparison of visual-only, text-only, and multimodal variants on GenVideo, measured by AP (\%), AUC (\%), and ACC (\%).}
    \resizebox{0.49\textwidth}{!}{
    \begin{tabular}{c | c c | c c c}
    \toprule
    Variant                      & Visual                     & \makecell{Caption-derived\\Text}           & AP              & AUC             & ACC            \\
    \midrule
    Visual-only                  & $\checkmark$               &                                            & 97.42           & 97.04           & 90.30          \\
    
    Text-only                    &                            & $\checkmark$                               & 93.18           & 92.81           & 85.64          \\
    
    \textbf{Multimodal} $G$      & $\checkmark$               & $\checkmark$                               & \textbf{98.81}  & \textbf{98.66}  & \textbf{93.80} \\
    
    \bottomrule
    \end{tabular}
    }
    \label{tab:motivation_multimodal}
\end{table}

\begin{figure*}[!t]
\centering

\subfloat[Three Temporal Characterization Levels]{
    \includegraphics[width=0.22\textwidth]
    {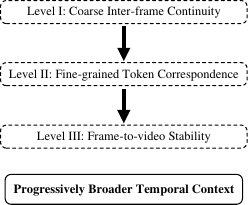}
    \label{fig:motivation_tor_a}
}
\hfill
\subfloat[Coarse Inter-frame Continuity]{
    \includegraphics[width=0.24\textwidth]
    {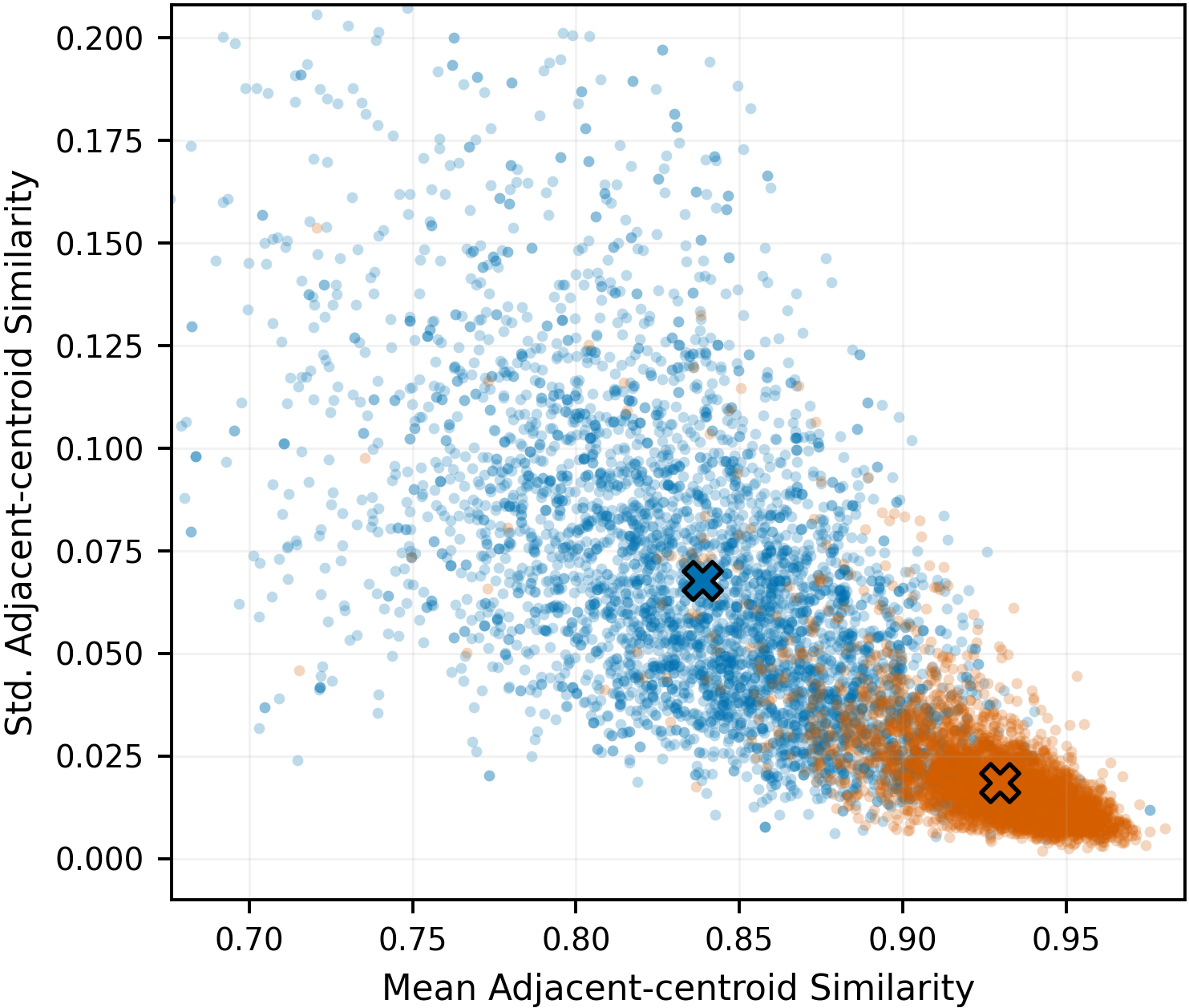}
    \label{fig:motivation_tor_b}
}
\hfill
\subfloat[Fine-grained Token Correspondence]{
    \includegraphics[width=0.245\textwidth]
    {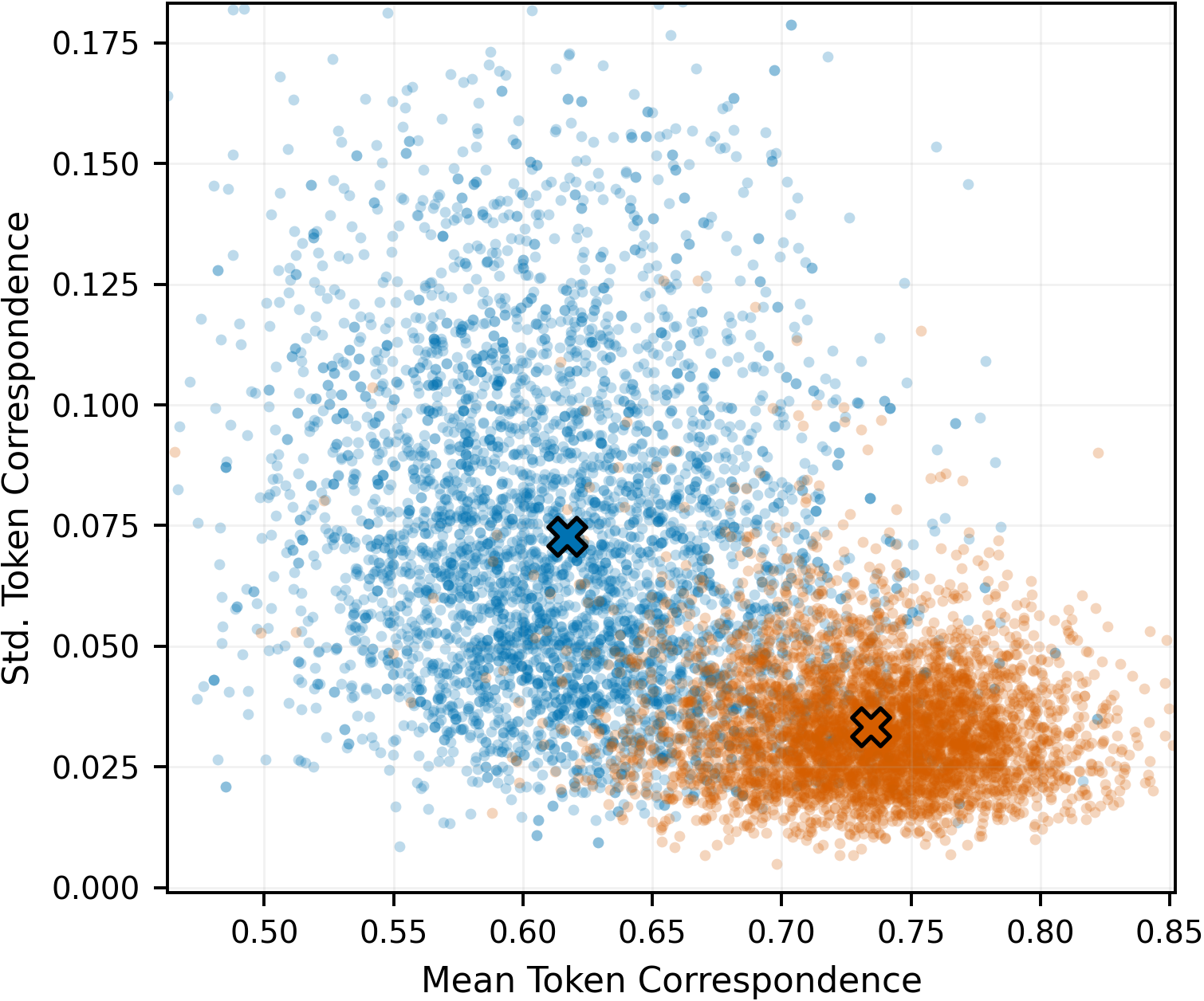}
    \label{fig:motivation_tor_c}
}
\hfill
\subfloat[Frame-to-video Stability]{
    \includegraphics[width=0.24\textwidth]
    {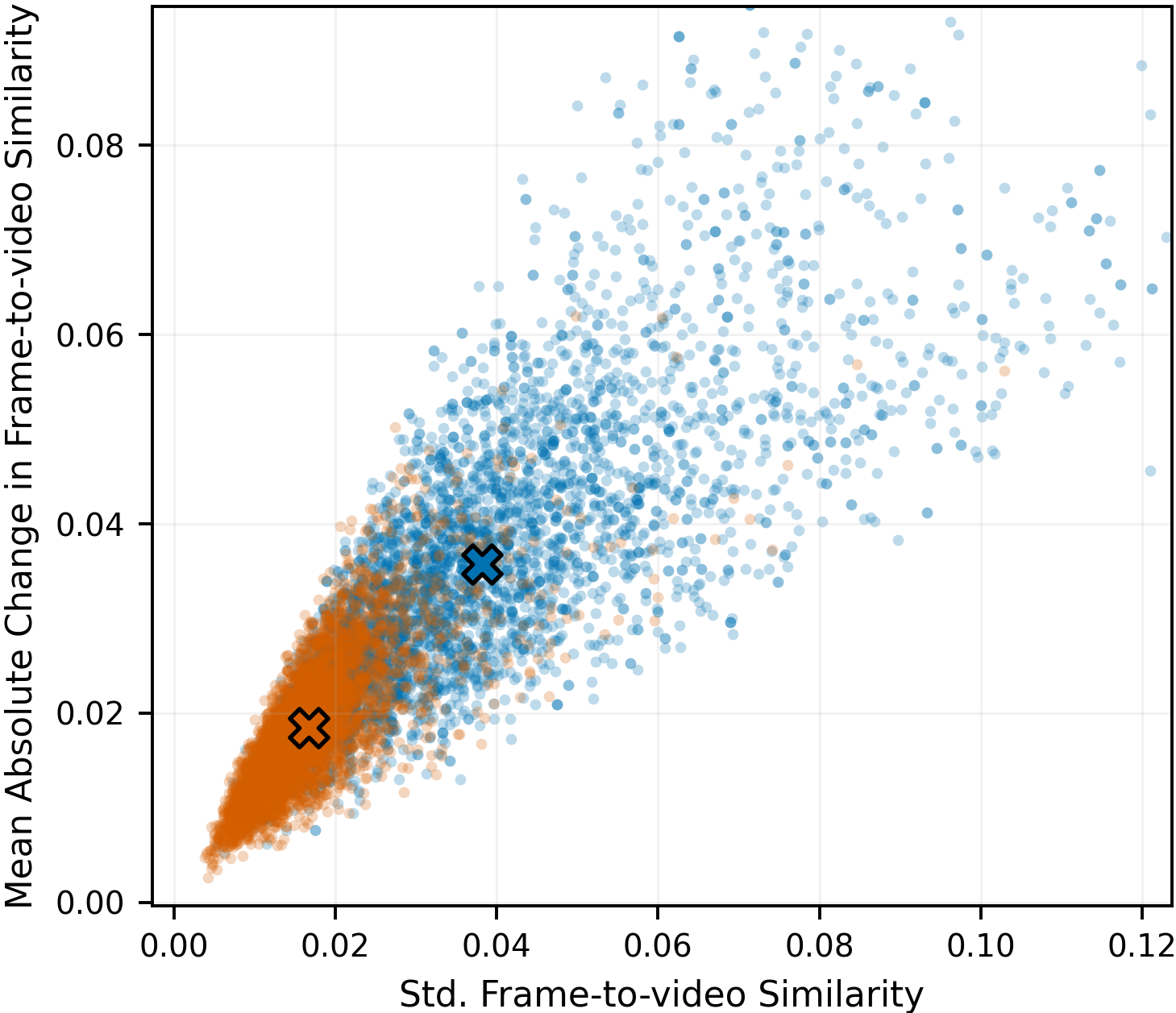}
    \label{fig:motivation_tor_d}
}
\caption{Temporal over-regularity on GenVideo. (a) Three levels of temporal characterization. (b) Coarse inter-frame continuity, visualized by the temporal mean and standard deviation of adjacent-centroid similarity. (c) Fine-grained token correspondence, visualized by the temporal mean and standard deviation of adjacent-token correspondence. (d) Frame-to-video stability, visualized by the standard deviation and mean absolute temporal change of frame-to-video similarity. Each point in (b)--(d) represents one video, with real and AI-generated videos shown in blue and orange, respectively; cross markers indicate the corresponding class means. Across all three levels, AI-generated videos consistently exhibit stronger temporal persistence and reduced temporal variability than real videos.}
\label{fig:motivation_tor}
\end{figure*}

\section{Empirical Observations and Motivation}
\label{sec:motivation}

This section presents the empirical observations underlying the design of MTOR.
We first examine video-level visual and caption-derived textual representations for AI-generated video detection and show that combining the two improves detection over either representation alone.
We then analyze temporal dynamics reflected in fine-grained visual representations and observe that AI-generated videos exhibit stronger temporal persistence and reduced temporal variability, a tendency we term \emph{temporal over-regularity} (TOR).
Finally, experiments across five benchmarks show that TOR complements the video-level multimodal representation, motivating the joint modeling of multimodal semantics and temporal over-regularity in MTOR.

\subsection{Video-Level Multimodal Representation}
\label{sec:motivation_multimodal}

To examine the effectiveness of different video-level representations for AI-generated video detection, we evaluate three variants: visual-only, text-only, and a multimodal variant $G$ that combines the two representations.
Given an input video \(\mathcal{V}\), a pretrained video encoder extracts its global visual representation, which is used by the visual-only variant for authenticity classification.
As reported in Table~\ref{tab:motivation_multimodal}, this variant achieves 97.42\% AP and 97.04\% AUC on GenVideo.

We next examine the caption-derived textual representation.
A video captioning model generates a textual description from the same input video, which is subsequently encoded into a video-level textual representation.
Since the caption is generated directly from $\mathcal{V}$, no external textual input is required.
As shown in Table~\ref{tab:motivation_multimodal}, the text-only variant achieves $93.18\%$ AP and $92.81\%$ AUC.

Although both representations are derived from the same video, they encode the video from different perspectives.
The visual representation encodes global spatiotemporal information, whereas the caption-derived textual representation summarizes high-level semantics such as objects, actions, attributes, and their relations.
Combining the two increases AP from $97.42\%$ to $98.81\%$, AUC from $97.04\%$ to $98.66\%$, and ACC from $90.30\%$ to $93.80\%$.
These gains indicate that caption-derived textual representations provide complementary discriminative cues to global visual representations.
Accordingly, MTOR jointly models the two representations in its video-level multimodal branch $G$, while fine-grained temporal dynamics are investigated next.

\subsection{Temporal Over-Regularity}
\label{sec:motivation_tor}

To investigate temporal regularity in fine-grained visual representations, we analyze visual tokens at three levels, as illustrated in Fig.~\ref{fig:motivation_tor}(a): coarse inter-frame continuity, fine-grained token correspondence, and frame-to-video stability.
Each level is characterized by statistics reflecting temporal persistence and variability.
Fig.~\ref{fig:motivation_tor}(b)--(d) visualize the distributions of real and AI-generated videos under these analyses.
Across all three levels, AI-generated videos exhibit stronger temporal persistence and reduced temporal variability than real videos.

\emph{Coarse Inter-frame Continuity:} 
We average the visual tokens within each frame to obtain a frame centroid and compute the cosine similarity between consecutive frame centroids.
Fig.~\ref{fig:motivation_tor}(b) plots the temporal mean of adjacent-centroid similarity against its standard deviation.
AI-generated videos exhibit higher mean similarity and lower temporal variation, indicating stronger inter-frame continuity.

\emph{Fine-grained Token Correspondence:}
We compare visual tokens between consecutive frames to characterize fine-grained temporal correspondence.
For each pair of adjacent frames, each token is matched to its most similar counterpart in the other frame, and the resulting similarities are aggregated into an adjacent-token correspondence score.
Fig.~\ref{fig:motivation_tor}(c) shows that AI-generated videos exhibit higher mean correspondence and lower temporal variation, indicating stronger and more stable token correspondence across consecutive frames.

\emph{Frame-to-video Stability:}
To characterize temporal regularity over a broader context, we compare the visual tokens of each frame with those from the remaining frames in the video, yielding a frame-to-video similarity.
Fig.~\ref{fig:motivation_tor}(d) characterizes the temporal variation of these similarities.
AI-generated videos exhibit lower variation and smaller temporal changes, indicating more stable frame-to-video relations over time.

Collectively, these three analyses reveal a consistent temporal tendency in AI-generated videos, characterized by stronger temporal persistence and reduced temporal variability.
We term this tendency \emph{temporal over-regularity} (TOR), which motivates the TOR component for AI-generated video detection.

\subsection{Complementarity of Multimodal Semantics and Temporal Over-Regularity}
\label{sec:motivation_complementarity}

\begin{table}[!ht]
\centering
\caption{Mean performance comparison of the multimodal branch $G$, the TOR-only variant, and MTOR across five benchmarks, measured by AP (\%), AUC (\%), and ACC (\%).}
\resizebox{0.3\textwidth}{!}{
\begin{tabular}{c | c c c}
\toprule
Variant          & AP             & AUC            & ACC            \\
\midrule
Multimodal $G$   & 97.26          & 97.17          & 89.51          \\
TOR-only         & 93.22          & 94.59          & 84.19          \\
\textbf{MTOR}    & \textbf{97.76} & \textbf{97.92} & \textbf{90.24} \\
\bottomrule
\end{tabular}
}
\label{tab:motivation_Complementarity}
\end{table}

The preceding analyses reveal two distinct characteristics of the video.
The multimodal representation captures global visual and caption-derived semantics, whereas TOR characterizes temporal over-regularity reflected in fine-grained visual dynamics.
We therefore examine whether the two provide complementary cues for AI-generated video detection.

Table~\ref{tab:motivation_Complementarity} reports the mean performance across five benchmarks: GenVideo, EvalCrafter, VideoPhy, VidProM, and GenVidBench.
The multimodal branch $G$ achieves 97.26\% AP, 97.17\% AUC, and 89.51\% ACC, while the TOR-only variant achieves 93.22\% AP, 94.59\% AUC, and 84.19\% ACC.
By combining these two predictions, MTOR improves performance to 97.76\% AP, 97.92\% AUC, and 90.24\% ACC.
These gains indicate that temporal over-regularity provides discriminative cues beyond the global multimodal representation.

These results demonstrate the complementarity between video-level multimodal semantics and temporal over-regularity.
Accordingly, MTOR combines the predictions from the multimodal branch and the TOR component through score-level fusion, as described in Sec.~\ref{sec:methodology}.

\begin{figure*}[t]
\centering
\includegraphics[width=\textwidth]{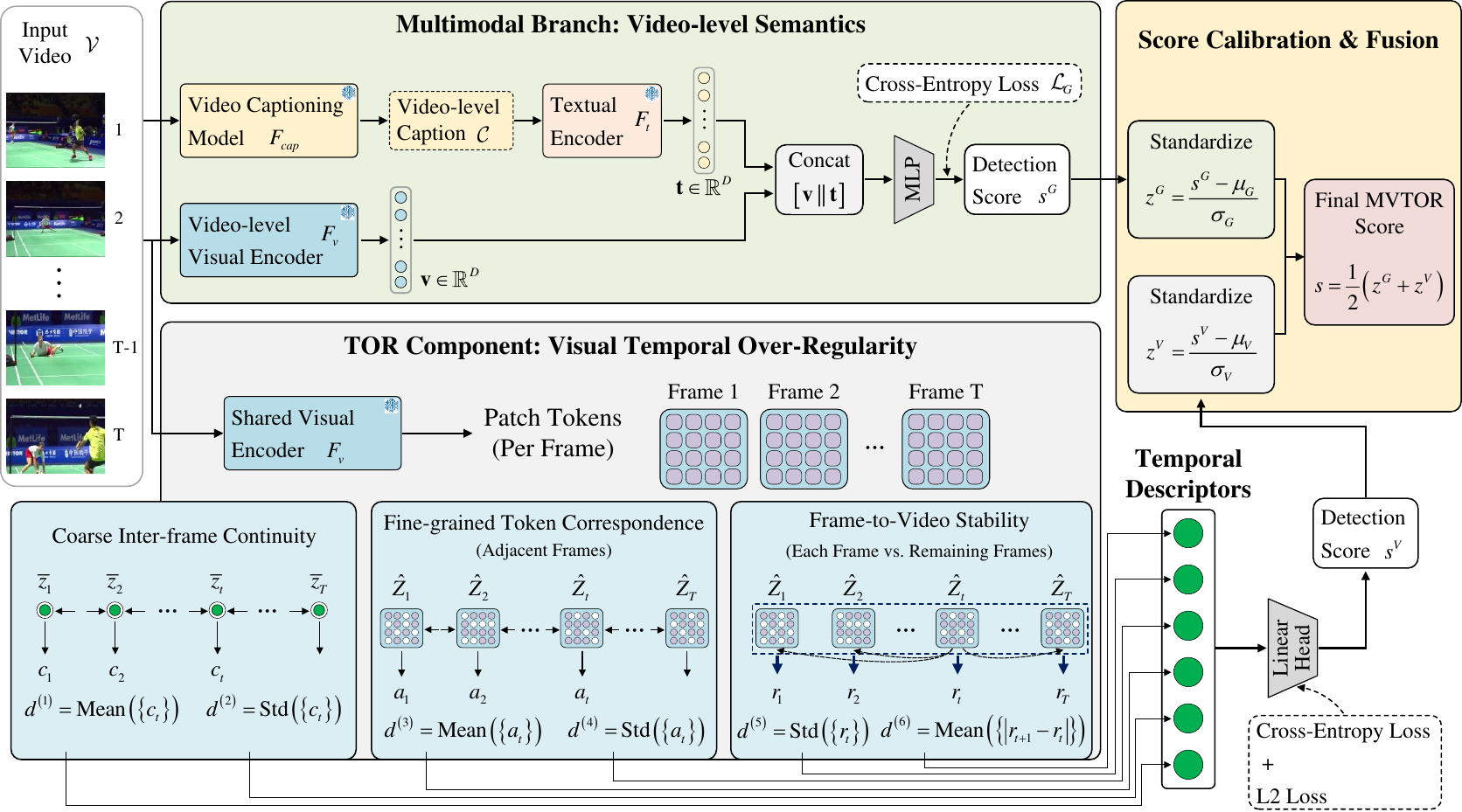} 
\caption{Overview of the proposed MTOR framework, consisting of the video-level multimodal branch, the TOR component, and source-calibrated score fusion.}
\label{fig2_MTOR}
\end{figure*}

\section{Methodology}
\label{sec:methodology}

Motivated by the empirical observations in Sec.~\ref{sec:motivation}, MTOR jointly models video-level multimodal semantics and temporal over-regularity for AI-generated video detection. 
The multimodal branch integrates global visual and caption-derived textual representations, whereas the TOR component models temporal over-regularity through coarse inter-frame continuity, fine-grained token correspondence, and frame-to-video stability. 
The two components produce complementary detection scores, which are calibrated using source-training statistics and subsequently fused to obtain the final MTOR score.

\subsection{Video-Level Multimodal Branch}
\label{sec:method_multimodal}

Given an input video $\mathcal{V}$, we employ a video encoder $F_v$, a video captioning model $F_{\mathrm{cap}}$, and a text encoder $F_t$, all of which remain frozen during training.
The video encoder produces a global visual representation:
\begin{equation}
    \mathbf{v}
    =
    F_v(\mathcal{V}),
    \qquad
    \mathbf{v}\in\mathbb{R}^{D}.
    \label{eq:global_visual}
\end{equation}
Meanwhile, the captioning model generates a textual description for the input video:

\begin{equation}
    \mathcal{C}
    =
    F_{\mathrm{cap}}(\mathcal{V}),
    \label{eq:video_caption}
\end{equation}
The caption is generated solely from the input video and does not require an externally provided text prompt.
The resulting caption $\mathcal{C}$ is then encoded into a global textual representation:
\begin{equation}
    \mathbf{t}
    =
    F_t(\mathcal{C}),
    \qquad
    \mathbf{t}\in\mathbb{R}^{D}.
    \label{eq:global_text}
\end{equation}

The global visual and textual representations are concatenated to form the multimodal representation:
\begin{equation}
    \mathbf{h}^{G}
    =
    \left[
        \mathbf{v}
        \,\Vert\,
        \mathbf{t}
    \right]
    \in
    \mathbb{R}^{2D},
    \label{eq:multimodal_representation}
\end{equation}
where $\Vert$ denotes feature concatenation.
A multilayer perceptron $g_{\theta}$ maps $\mathbf{h}^{G}$ to two classification logits:
\begin{equation}
    \boldsymbol{\ell}^{G}
    =
    g_{\theta}
    \left(
        \mathbf{h}^{G}
    \right)
    =
    \left[
        \ell_{0}^{G},
        \ell_{1}^{G}
    \right],
    \label{eq:global_logits}
\end{equation}
where $\ell_{0}^{G}$ and $\ell_{1}^{G}$ correspond to the real and AI-generated classes, respectively. We define the multimodal detection score as the logit margin:
\begin{equation}
    s^{G}
    =
    \ell_{1}^{G}
    -
    \ell_{0}^{G}.
    \label{eq:global_score}
\end{equation}

For the $n$-th video $\mathcal{V}_n$ in the source-training set, let $\boldsymbol{\ell}_n^G = [\ell_{n,0}^G,\ell_{n,1}^G]$ denote the corresponding logits, and $y_n\in\{0,1\}$ denote its label.
The multimodal prediction head is optimized using the cross-entropy loss:
\begin{equation}
    \mathcal{L}_{G}
    =
    -\frac{1}{N}
    \sum_{n=1}^{N}
    \sum_{c=0}^{1}
    \mathbb{I}(y_n=c)
    \log
    \frac{
        \exp(\ell_{n,c}^{G})
    }{
        \sum_{j=0}^{1}
        \exp(\ell_{n,j}^{G})
    },
    \label{eq:multimodal_loss}
\end{equation}
where $\mathbb{I}(\cdot)$ denotes the indicator function.

\subsection{Temporal Over-Regularity Modeling}
\label{sec:method_tor}

The TOR component characterizes temporal regularity using fine-grained visual tokens produced by the frozen video encoder.
For a video represented by $T$ sampled frames, let
\begin{equation}
    \mathbf{Z}_{t}
    =
    \left[
        \mathbf{z}_{t,1},
        \mathbf{z}_{t,2},
        \ldots,
        \mathbf{z}_{t,K}
    \right]
    \in
    \mathbb{R}^{K\times d},
    \qquad
    t=1,\ldots,T,
    \label{eq:visual_tokens}
\end{equation}
denote the $K$ visual tokens associated with frame $t$, where each token is a learned representation associated with a local image region.
For token-level comparison across frames, each token is $\ell_2$-normalized as:
\begin{equation}
    \hat{\mathbf{z}}_{t,k}
    =
    \frac{
        \mathbf{z}_{t,k}
    }{
        \|\mathbf{z}_{t,k}\|_2
    },
    \qquad
    k=1,\ldots,K.
    \label{eq:token_normalization}
\end{equation}
We denote the resulting normalized token set by:
\begin{equation}
    \hat{\mathbf{Z}}_{t}
    =
    \left[
        \hat{\mathbf{z}}_{t,1},
        \ldots,
        \hat{\mathbf{z}}_{t,K}
    \right]^{\top}.
\end{equation}
TOR characterizes temporal over-regularity through three levels: coarse inter-frame continuity, fine-grained token correspondence, and frame-to-video stability.

\noindent\textbf{Coarse Inter-frame Continuity.}
We average the visual tokens within each frame and normalize the resulting frame centroid:
\begin{equation}
    \bar{\mathbf{z}}_{t}
    =
    \frac{
        \frac{1}{K}
        \sum_{k=1}^{K}
        \mathbf{z}_{t,k}
    }{
        \left\|
        \frac{1}{K}
        \sum_{k=1}^{K}
        \mathbf{z}_{t,k}
        \right\|_2
    }.
    \label{eq:frame_centroid}
\end{equation}
The cosine similarity between consecutive frame centroids is then computed as:
\begin{equation}
    c_t
    =
    \bar{\mathbf{z}}_{t}^{\top}
    \bar{\mathbf{z}}_{t+1},
    \qquad
    t=1,\ldots,T-1.
    \label{eq:centroid_similarity}
\end{equation}
The temporal mean and standard deviation of \(c_t\) characterize the strength and variation of coarse inter-frame continuity:
\begin{equation}
    d^{(1)}
    =
    \frac{1}{T-1}
    \sum_{t=1}^{T-1}
    c_t,
    \qquad
    d^{(2)}
    =
    \operatorname{Std}
    \left(
        \{c_t\}_{t=1}^{T-1}
    \right),
    \label{eq:centroid_descriptors}
\end{equation}
where $\operatorname{Std}(\cdot)$ denotes the standard deviation.

\noindent\textbf{Fine-grained Token Correspondence.}
To characterize fine-grained token correspondence, we match visual tokens between consecutive frames.
For two sets of normalized visual tokens
$\mathcal{A}=\{\mathbf{a}_p\}_{p=1}^{P}$
and
$\mathcal{B}=\{\mathbf{b}_q\}_{q=1}^{Q}$,
their bidirectional correspondence strength is defined as:
\begin{equation}
\begin{aligned}
    \mathcal{M}(\mathcal{A},\mathcal{B})
    =
    \frac{1}{2}
    \Bigg[
    &\frac{1}{P}
    \sum_{p=1}^{P}
    \max_{1\leq q\leq Q}
    \mathbf{a}_{p}^{\top}\mathbf{b}_{q}
    \\
    +
    &\frac{1}{Q}
    \sum_{q=1}^{Q}
    \max_{1\leq p\leq P}
    \mathbf{b}_{q}^{\top}\mathbf{a}_{p}
    \Bigg].
    \label{eq:bidirectional_matching}
\end{aligned}
\end{equation}
For each token in one set, the maximum similarity identifies its strongest correspondence in the other set. Averaging over both directions makes the measure symmetric with respect to the two token sets.
For each pair of consecutive frames, the adjacent-token correspondence score is computed as:
\begin{equation}
    a_t
    =
    \mathcal{M}
    \left(
        \hat{\mathbf{Z}}_{t},
        \hat{\mathbf{Z}}_{t+1}
    \right),
    \qquad
    t=1,\ldots,T-1.
    \label{eq:adjacent_token_similarity}
\end{equation}
The temporal mean and standard deviation of \(a_t\) characterize fine-grained token correspondence in terms of its strength and temporal variation over time:
\begin{equation}
    d^{(3)}
    =
    \frac{1}{T-1}
    \sum_{t=1}^{T-1}
    a_t,
    \qquad
    d^{(4)}
    =
    \operatorname{Std}
    \left(
        \{a_t\}_{t=1}^{T-1}
    \right).
    \label{eq:adjacent_token_descriptors}
\end{equation}

\noindent\textbf{Frame-to-video Stability.}
The preceding two aspects focus on temporal consistency between consecutive frames.
To characterize temporal regularity over broader contexts, we compare each frame against the remaining video context.
For frame $t$, the normalized visual tokens from all remaining frames are concatenated to form the corresponding video context:
\begin{equation}
    \mathbf{R}_{t}
    =
    \operatorname{Concat}
    \left(
        \hat{\mathbf{Z}}_{1},
        \ldots,
        \hat{\mathbf{Z}}_{t-1},
        \hat{\mathbf{Z}}_{t+1},
        \ldots,
        \hat{\mathbf{Z}}_{T}
    \right)
    \in
    \mathbb{R}^{(T-1)K\times d}.
    \label{eq:remaining_token_set}
\end{equation}
The frame-to-video similarity is defined as:
\begin{equation}
    r_t
    =
    \mathcal{M}
    \left(
        \hat{\mathbf{Z}}_{t},
        \mathbf{R}_{t}
    \right),
    \qquad
    t=1,\ldots,T.
    \label{eq:video_context_similarity}
\end{equation}
Frame-to-video stability is characterized by the temporal variability of \(r_t\).
We quantify this stability using the standard deviation across frames:
\begin{equation}
    d^{(5)}
    =
    \operatorname{Std}
    \left(
        \{r_t\}_{t=1}^{T}
    \right),
    \label{eq:context_std}
\end{equation}
and the mean absolute change between consecutive frames:
\begin{equation}
    d^{(6)}
    =
    \frac{1}{T-1}
    \sum_{t=1}^{T-1}
    \left|
        r_{t+1}
        -
        r_t
    \right|.
    \label{eq:context_change}
\end{equation}

The six temporal descriptors are collected into:
\begin{equation}
    \mathbf{d}
    =
    \left[
        d^{(1)},
        d^{(2)},
        d^{(3)},
        d^{(4)},
        d^{(5)},
        d^{(6)}
    \right]^{\top}
    \in
    \mathbb{R}^{6}.
    \label{eq:tor_descriptor}
\end{equation}
Together, the six descriptors characterize temporal persistence and variability through coarse inter-frame continuity, fine-grained token correspondence, and frame-to-video stability.
To place the six descriptors on comparable scales, each dimension is standardized using training-set statistics:
\begin{equation}
    \tilde{d}^{(m)}
    =
    \frac{
        d^{(m)}
        -
        \mu_{m}^{d}
    }{
        \sigma_{m}^{d}
    },
    \qquad
    m=1,\ldots,6,
    \label{eq:descriptor_standardization}
\end{equation}
where \(\mu_m^d\) and \(\sigma_m^d\) denote the corresponding mean and standard deviation estimated on the source-training set. These statistics are fixed thereafter.
We then form the standardized temporal descriptor vector as:
\begin{equation}
    \tilde{\mathbf{d}}
    =
    \left[
        \tilde{d}^{(1)},
        \ldots,
        \tilde{d}^{(6)}
    \right]^{\top}.
\end{equation}

A bias-free linear head maps the standardized temporal descriptors to the detection score:
\begin{equation}
    s^{V}
    =
    \mathbf{w}^{\top}
    \tilde{\mathbf{d}},
    \qquad
    \mathbf{w}\in\mathbb{R}^{6}.
    \label{eq:tor_score}
\end{equation}
For training video \(V_n\), let \(s_n^V\) denote its TOR score.
The TOR linear head is optimized using a binary cross-entropy objective together with $\ell_2$ regularization:
\begin{equation}
\begin{aligned}
    \mathcal{L}_{V}
    =
    &-\frac{1}{N}
    \sum_{n=1}^{N}
    \left[
        y_n\log\sigma(s_n^{V})
        +
        (1-y_n)
        \log
        \left(
            1-\sigma(s_n^{V})
        \right)
    \right]
    \\
    &+
    \lambda
    \|\mathbf{w}\|_2^2,
    \label{eq:tor_loss}
\end{aligned}
\end{equation}
where $y_n\in\{0,1\}$ denotes the ground-truth label and $\lambda$ controls the $\ell_2$ regularization.

\subsection{Source-Calibrated Score Fusion}
\label{sec:score_fusion}

The multimodal and TOR scores are first calibrated to a common scale. Let \(\mu_G\) and \(\sigma_G\) denote the mean and standard deviation of the multimodal scores, and \(\mu_V\) and \(\sigma_V\) those of the TOR scores, estimated on the source-training set. The standardized scores are defined as:
\begin{equation}
    z^{G}
    =
    \frac{
        s^{G}
        -
        \mu_G
    }{
        \sigma_G
    },
    \qquad
    z^{V}
    =
    \frac{
        s^{V}
        -
        \mu_V
    }{
        \sigma_V
    }.
    \label{eq:score_standardization}
\end{equation}
They are fused by averaging to obtain the final MTOR score:
\begin{equation}
    s
    =
    \frac{1}{2}
    \left(
        z^{G}
        +
        z^{V}
    \right).
    \label{eq:mtor_fusion}
\end{equation}
The calibration statistics are fixed after training, and no target-domain samples are used during inference.

\section{Experiment}

\subsection{Datasets}
We evaluate MTOR on five benchmarks covering diverse video generators and data distributions, including GenVideo~\cite{chen2024demamba}, EvalCrafter~\cite{liu2024evalcrafter}, VideoPhy~\cite{bansal2024videophy}, VidProM~\cite{wang2024vidprom}, and GenVidBench~\cite{Ni2025GenVidBenchA6}.

\noindent\textbf{Training Set.}
We use the official training split of GenVideo as the source dataset and further divide it into source-training and source-validation subsets. 
Real videos are sampled from the Youku-mPLUG dataset~\cite{xu2023youku}, and synthetic videos are generated by Pika~\cite{pika2022}.

\noindent\textbf{Validation Set.}
We randomly reserve 20\% of the source data as a stratified validation set for learning-rate scheduling, early stopping, and hyperparameter selection.

\noindent\textbf{Testing Set.}
The testing set collects synthetic videos from 46 generator variants across five benchmarks, with real videos uniformly sampled from MSR-VTT~\cite{xu2016msr} following GenVideo’s standard setting.
\begin{itemize}

\item \textbf{GenVideo}~\cite{chen2024demamba}: Comprises 10 distinct test subsets spanning a diverse range of generative architectures, specifically featuring Gen2~\cite{esser2023structure}, Sora~\cite{brooks2024video}, ModelScope (MS)~\cite{wang2023modelscope}, MoonValley (MV)~\cite{moonvalley2022}, MorphStudio (MPS)~\cite{morph2023}, Show-1~\cite{zhang2024show}, HotShot~\cite{hotshot2023}, LaVie~\cite{wang2025lavie}, Crafter~\cite{chen2023videocrafter1}, and WildScrape (WS)~\cite{wei2024dreamvideo, feng2023dreamoving, xu2024magicanimate}.

\item \textbf{EvalCrafter}~\cite{liu2024evalcrafter}: Features 14 video generator subsets, namely VideoCrafter V0.9 (Floor33), MoonValley (MV), Gen2, Gen2-December (Gen2-D), Mix-SR, HotShot-XL (HS-XL), Show-1, ModelScope (MS), LaVie-Base (LaVie-B), LaVie-Interpolation (LaVie-I), PikaLab (PKL), VideoCrafter (VC), PikaLab V1.0 (PKL-V1), and ZeroScope (ZS).

\item \textbf{VideoPhy}~\cite{bansal2024videophy}: Incorporates 10 specialized subsets tailored for evaluating physical world dynamics and spatiotemporal consistency, which are synthesized by CogVideoX (CVX), CogVideoX-5B (CVX-5B), Dream-Machine (DM), Gen2, LaVie, OpenSora, Pika, SVD-T2I2V, VideoCrafter2 (VC2), and ZeroScope (ZS).

\item \textbf{VidProM}~\cite{wang2024vidprom}: Encompasses 6 distinct text-to-video generation tracks derived from diverse real-world user prompts, including ModelScope (MS), OpenSora, Pika, StreamingT2V (ST2V), Text2Video-Zero (T2VZ), and VideoCrafter2 (VC2).

\item \textbf{GenVidBench}~\cite{Ni2025GenVidBenchA6}: Integrates 6 highly challenging generator sets representing the contemporary frontier of state-of-the-art video synthesis, generated by MuseV, SVD, Mora, CogVideo, Sora, and Kling.

\end{itemize}

\begin{table*}[th]
    \centering
    \caption{Comparison of AP (\%) between MTOR and 16 representative baselines on GenVideo. \textsuperscript{\textdagger} Results reproduced using the official code. \textsuperscript{\textdagger\textdagger} Results reproduced from our implementation because no official code is available.}
    \resizebox{\textwidth}{!}{
    \begin{tabular}{c | c c c c c c c c c c | c}
    \toprule
    \rowcolor{gray!25}
    Method                                                         & MS        & MPS        & MV         & HotShot            & Show-1            & Gen2               & Crafter               & LaVie               & Sora                 & WS             & mean                \\
    \midrule
    \hline
    \rowcolor{gray!5}  
    FID\textsuperscript{\textdagger} \cite{NEURIPS2024_6dddcff5}, NeurIPS 2024          & 91.50             & 92.24              & 93.67              & 86.10  & 90.61 & 93.27              & 92.41                 & 83.68               & 74.95                & 82.24      & 88.07   \\ 
    NPR\textsuperscript{\textdagger} \cite{Tan_2024_CVPR}, CVPR 2024                    & 84.67             & 96.53              & 96.79              & 40.17              & 21.61             & 96.35              & 97.02                 & 22.37               & 90.55    & 66.51                  & 71.26               \\
    \hline
    \rowcolor{gray!5}
    STIL\textsuperscript{\textdagger} \cite{10.1145/3474085.3475508}, ACM MM 2021       & 88.21             & 88.68              & 73.07              & 56.91              & 62.07             & 83.96              & 64.87                 & 65.54               & 49.86                & 63.60                  & 69.68               \\ 
    FTCN\textsuperscript{\textdagger} \cite{zheng2021exploring}, ICCV 2021              & 70.01             & 83.59              & 97.07              & 87.42              & 93.30             & 91.86              & 91.72                 & 84.16               & 44.48                & 84.46                  & 82.81               \\
    \rowcolor{gray!5}
    TALL\textsuperscript{\textdagger} \cite{xu2023tall}, ICCV 2023                      & 51.11             & 63.63              & 92.09              & 44.00              & 51.06             & 93.47              & 87.85                 & 59.07               & 15.82                & 64.43                  & 62.25               \\ 
    MINTIME\textsuperscript{\textdagger} \cite{10547206}, TIFS 2024                    & 79.27             & 82.03              & 89.80              & 87.68              & 89.23             & 88.26              & 87.34                 & 82.48               & 80.75                & 85.10                  & 85.19               \\
    \hline
    \rowcolor{gray!5}    
    AIGVDet\textsuperscript{\textdagger} \cite{bai2024ai}, PRCV 2024                    & 70.91             & 67.93              & 56.22              & 51.81              & 72.59             & 89.98              & 75.87                 & 88.62   & 65.70                & 64.96                  & 70.46               \\
    DeMamba\textsuperscript{\textdagger} \cite{chen2024demamba}, arXiv 2024             & 41.96             & 97.07              & 84.64              & 67.63              & 45.07             & 96.11              & \underline{98.26}                 & 81.49               & 28.79                & 78.00                  & 71.90               \\
    \rowcolor{gray!5} 
    DeCoF\textsuperscript{\textdagger\textdagger} \cite{Ma2024DetectingAV}, ICME 2025   & 91.18             & 91.69              & \underline{98.68}              & 76.02              & 48.99             & 98.28              & 94.67                 & 77.79               & 55.76                & 73.68                  & 81.67               \\
    NSG-VD\textsuperscript{\textdagger}~\cite{zhang2025NSGVD}, NeurIPS 2025        & 61.45 & 62.22 & 52.92 & 61.87 & 57.69 & 48.50 & 68.90 & 56.72 & 62.74 & 64.56 & 59.76  \\
    \rowcolor{gray!5}
    ReStraV\textsuperscript{\textdagger} \cite{interno2025aigenerated}, NeurIPS 2025    & \underline{95.34}             & \underline{97.22}              & 97.43              & \textbf{99.19}              & 74.52             & \textbf{99.92}             & 71.28             & \underline{96.80}             & \textbf{97.99}             & 73.75             & 90.34              \\
    D3\textsuperscript{\textdagger} \cite{Zheng_2025_ICCV}, ICCV 2025                   & 85.59             & 94.07             & 96.22             & 97.09             & 95.29             & 94.65             & 96.46             & 88.22             & 87.71             & 86.30             & \underline{92.16}         \\ 
    \rowcolor{gray!5}
    STALL\textsuperscript{\textdagger}  \cite{hayun2026stall}, CVPR 2026                & 77.75            & 83.60            & 75.54            & 77.79            & 80.20            & 88.40            & 78.02            & 84.22            & 73.46            & 66.43            & 78.54            \\ 
    \hline
    STPAG\textsuperscript{\textdagger\textdagger}  \cite{Rai_2024_CVPR}, CVPRW 2024   & 78.89 & 83.06 & 94.71 & 86.76 & 94.14 & 89.72 & 89.91 & 82.97 & 56.37 & 85.18 & 84.17      \\ 
    \rowcolor{gray!5} 
    LGN-Net\textsuperscript{\textdagger}  \cite{ZHAO2025125581}, ESWA 2025            & 88.40 & 89.38 & 97.11 & 96.37 & \underline{96.25} & 95.14 & 95.14 & 88.13 & 58.63 & \underline{91.36} & 89.59   \\ 
    X-CLIP\textsuperscript{\textdagger} \cite{ni2022expanding}, ECCV 2022               & 79.84             & 87.54              & 95.53              & 90.71              & 94.54             & 88.69              & 93.50                 & 86.28               & 64.23                & 88.54                  & 89.62               \\  
    \hline \hline
    \rowcolor{gray!25}
    \textbf{MTOR}       & \textbf{98.91}            & \textbf{99.22}            & \textbf{99.89}            & \underline{99.17}            & \textbf{98.79}            & \underline{99.84}            & \textbf{99.72}            & \textbf{98.26}            & \underline{96.59}            & \textbf{97.06}            & \textbf{98.74}            \\
    \bottomrule
    \end{tabular}
    }
\label{tab:ap_GenVideo}
\end{table*}
\begin{table*}[ht]
    \centering
    \caption{Comparison of AUC (\%) between MTOR and 16 representative baselines on GenVideo. \textsuperscript{\textdagger} Results reproduced using the official code. \textsuperscript{\textdagger\textdagger} Results reproduced from our implementation because no official code is available.}
    \resizebox{\textwidth}{!}{
    \begin{tabular}{c | c c c c c c c c c c | c}
    \toprule
    \rowcolor{gray!25}
    Method                                                                                     & MS     & MPS     & MV      & HotShot           & Show-1            & Gen2                 & Crafter                 & LaVie                 & Sora                 & WS                 & mean   \\
    \midrule
    \hline 
    \rowcolor{gray!5}
    FID\textsuperscript{\textdagger} \cite{NEURIPS2024_6dddcff5}, NeurIPS 2024                  & 90.94             & 91.93           & 93.70             & 85.77 & 91.19 & 93.16                & 92.02                   & 82.57     & 73.45                & 81.30          & 87.60  \\ 
    NPR\textsuperscript{\textdagger} \cite{Tan_2024_CVPR}, CVPR 2024                            & 93.92             & \textbf{99.38}  & \textbf{99.92}    & 26.83             & 18.99             & 98.78                & \underline{99.24}          & 42.32                 & \textbf{97.56}    & 76.12                      & 75.31  \\
    \hline
    \rowcolor{gray!5} 
    STIL\textsuperscript{\textdagger} \cite{10.1145/3474085.3475508}, ACM MM 2021               & 86.48             & 88.75           & 82.77             & 60.00             & 68.71             & 86.67                & 73.56                   & 72.67                 & 54.18                & 69.94                      & 74.37  \\
    FTCN\textsuperscript{\textdagger} \cite{zheng2021exploring}, ICCV 2021                      & 69.76             & 83.25           & 97.18           & 88.69             & 93.45             & 93.33                & 91.62                   & 84.70               & 38.30                & 85.74                  & 82.60     \\
    \rowcolor{gray!5}
    TALL\textsuperscript{\textdagger} \cite{xu2023tall}, ICCV 2023                              & 58.46             & 60.54           & 83.24             & 45.21             & 46.24             & 73.30                & 66.37                   & 48.40                 & 66.36                & 53.84                      & 60.20   \\
    MINTIME\textsuperscript{\textdagger} \cite{10547206}, TIFS 2024                            & 81.27             & 85.48              & 90.97              & 90.34              & 90.89             & 88.80              & 89.77                 & 83.86               & 79.27                & 85.71                  & 86.64               \\ 
    \hline
    \rowcolor{gray!5} 
    AIGVDet\textsuperscript{\textdagger} \cite{bai2024ai}, PRCV 2024                            & 68.05             & 79.42           & 59.41             & 74.67             & 70.29             & 71.62                & 69.77                   & 79.55                 & 60.79                & 67.82                      & 70.14  \\ 
    DeMamba\textsuperscript{\textdagger} \cite{chen2024demamba}, arXiv 2024                     & 80.53             & 95.52           & \underline{99.90} & 59.49             & 49.34             & 99.10    & 97.99                   & 63.25                 & 91.76                & 71.15                      & 80.80  \\ 
    \rowcolor{gray!5} 
    DeCoF\textsuperscript{\textdagger\textdagger} \cite{Ma2024DetectingAV}, ICME 2025           & 95.49             & 96.87           & 99.60           & 90.22             & 77.86             & 98.94                & 96.63                   & 84.62                 & 96.36                & 87.72                      & 92.43   \\
    NSG-VD\textsuperscript{\textdagger}~\cite{zhang2025NSGVD}, NeurIPS 2025      & 71.56 & 72.74 & 61.30 & 73.49 & 70.96 & 57.15 & 78.07 & 66.57 & 76.41 & 70.26 & 69.85  \\
    \rowcolor{gray!5}
    ReStraV\textsuperscript{\textdagger} \cite{interno2025aigenerated}, NeurIPS 2025            & \underline{96.84}             & 98.20              & 96.95              & \textbf{99.27}              & 79.65             & \textbf{99.93}             & 77.40             & \underline{97.75}             & \underline{97.35}             & 81.55             & \underline{92.49}              \\
    D3\textsuperscript{\textdagger} \cite{Zheng_2025_ICCV}, ICCV 2025                           & 86.00             & 93.62           & 95.74           & 96.96           & 94.03           & 93.93           & 96.21           & 88.93           & 88.93           & 87.60           & 92.20       \\  
    \rowcolor{gray!5} 
    STALL\textsuperscript{\textdagger}  \cite{hayun2026stall}, CVPR 2026                        & 77.64             & 82.60           & 72.03           & 78.79           & 81.79           & 87.32           & 80.05           & 84.73           & 70.50           & 72.32           & 78.78            \\ 
    \hline
    STPAG\textsuperscript{\textdagger\textdagger}  \cite{Rai_2024_CVPR}, CVPRW 2024   & 79.46 & 83.59 & 95.42 & 88.86 & 95.31 & 92.06 & 90.77 & 83.20 & 56.09 & 87.13 & 85.19 \\ 
    \rowcolor{gray!5} 
    LGN-Net\textsuperscript{\textdagger}  \cite{ZHAO2025125581}, ESWA 2025            & 88.07 & 88.72 & 97.16 & 97.34 & \underline{96.91} & 96.43 & 94.65 & 88.48 & 54.97 & \underline{92.30} & 89.50 \\ 
    X-CLIP\textsuperscript{\textdagger} \cite{ni2022expanding}, ECCV 2022                       & 80.51             & 86.14           & 95.12           & 91.51             & 94.69             & 88.96                & 92.84                   & 85.04               & 64.09                & 87.93                  & 88.87     \\  
    \hline \hline
    \rowcolor{gray!25}
    \textbf{MTOR}       & \textbf{98.90}            & \underline{99.24}            & 99.89            & \underline{99.21}            & \textbf{98.80}            & \underline{99.85}            & \textbf{99.74}            & \textbf{98.22}            & 95.60            & \textbf{96.92}            & \textbf{98.64}            \\
    \bottomrule
    \end{tabular}
    }
\label{tab:auc_GenVideo}
\end{table*}

\subsection{Implementation Details}
We employ ``ShareGPT4Video-8B''~\cite{chen2024sharegptvideo} to generate video captions and ``InternVideo2.5-Chat-8B''~\cite{wang2025internvideo} to extract visual and textual representations.
For each video, $T=8$ frames are uniformly sampled and resized to $448 \times 448$.
The global visual and caption-derived textual representations have dimension $D=4096$ and are $L_2$-normalized before concatenation.
For the TOR component, $K=16$ fine-grained visual tokens with dimension $d=4096$ are extracted from each sampled frame.

The multimodal MLP uses hidden dimensions of $2048$, $512$, and $128$ with a dropout rate of $0.5$.
It is trained using AdamW~\cite{loshchilov2018decoupled} with a learning rate of $1\times10^{-4}$, a weight decay of $1\times10^{-5}$, and a batch size of $64$ for up to $50$ epochs.
Early stopping uses a patience of $10$ epochs, while ``ReduceLROnPlateau'' reduces the learning rate by a factor of $0.5$ with a patience of $5$ based on source-validation AUC.
The six temporal descriptors are mapped to the TOR score by a bias-free linear head optimized with binary cross-entropy and $\ell_2$ regularization.
The regularization coefficient is set to $ \lambda=10^{-3} $, as selected on the source-validation set.

\begin{table*}[th]
    \centering
    \caption{Comparison of AP (\%) between MTOR and 16 representative baselines on EvalCrafter. \textsuperscript{\textdagger} Results reproduced using the official code. \textsuperscript{\textdagger\textdagger} Results reproduced from our implementation because no official code is available.}
    \resizebox{\textwidth}{!}{
    \begin{tabular}{c | c c c c c c c c c c c c c c| c}
    \toprule
    \rowcolor{gray!25}
    Method &  \makecell[c]{\rotatebox{50}{Floor33}} & \makecell[c]{\rotatebox{50}{Gen2}} &  \makecell[c]{\rotatebox{50}{Gen2-D}} &  \makecell[c]{\rotatebox{50}{HS-XL}} &  \makecell[c]{\rotatebox{50}{LaVie-B}} &  \makecell[c]{\rotatebox{50}{LaVie-I}} &  \makecell[c]{\rotatebox{50}{Mix-SR}} &  \makecell[c]{\rotatebox{50}{MS}} &  \makecell[c]{\rotatebox{50}{MV}} &  \makecell[c]{\rotatebox{50}{PKL}} &  \makecell[c]{\rotatebox{50}{PKL-V1}} &  \makecell[c]{\rotatebox{50}{Show-1}} &  \makecell[c]{\rotatebox{50}{VC}} & \makecell[c]{\rotatebox{50}{ZS}} &  \makecell[c]{\rotatebox{50}{mean}} \\
    \midrule
    \hline
    \rowcolor{gray!5} 
    FID\textsuperscript{\textdagger}                & 96.40        & 97.36        & 98.68        & 89.90        & 92.92        & 84.19        & 98.51        & 95.74        & 98.29        & 99.49        & 99.17        & \underline{96.77}        & 95.71        & 95.18        & \underline{95.59}        \\ 
    NPR\textsuperscript{\textdagger}                & \textbf{99.77}        & 99.34        & \textbf{99.95}        & 47.39        & 76.45        & 72.23        & \underline{99.67}   & \underline{98.54}        & \textbf{99.96}        & \underline{99.97}        & \textbf{99.93}        & 69.82        & \textbf{99.68}        & \underline{98.21}        & 90.07        \\
    \hline
    \rowcolor{gray!5}
    STIL\textsuperscript{\textdagger}               & 90.42        & 90.59        & 68.07        & 59.36        & 65.96        & 66.02        & 65.41        & 89.20        & 74.71        & 95.86        & 79.97        & 60.21        & 67.14        & 89.73        & 75.90        \\
    FTCN\textsuperscript{\textdagger}               & 82.85        & 88.50        & 94.72        & 88.25        & 82.71        & 81.91        & 92.74        & 71.16        & 96.65        & 96.36        & 93.14        & 93.55        & 92.23        & 69.09        & 87.42        \\
    \rowcolor{gray!5} 
    TALL\textsuperscript{\textdagger}               & 63.25        & 70.75        & 77.04        & 46.93        & 52.87        & 52.53        & 78.16        & 62.11        & 83.63        & 65.33        & 70.98        & 48.00        & 60.50        & 51.73        & 63.13        \\ 
    MINTIME\textsuperscript{\textdagger}           & 84.62        & 86.20        & 88.02        & 90.07        & 83.12        & 82.99        & 88.87        & 78.47        & 90.56        & 94.65        & 91.31        & 87.99         & 88.18        & 88.43        & 87.39        \\  
    \hline
    \rowcolor{gray!5}  
    AIGVDet\textsuperscript{\textdagger}            & 67.84        & 71.86        & 74.24        & 51.46        & 73.81        & 70.72        & 57.64        & 71.00        & 56.50        & 94.95        & 92.92        & 72.41        & 64.58        & 67.00        & 70.50        \\ 
    DeMamba\textsuperscript{\textdagger}            & 97.50        & 89.82        & 97.67        & 66.31        & 75.37        & 69.51        & 96.38        & 41.13        & 85.79        & 70.08        & 34.32        & 46.13        & 98.19        & 96.34        & 76.04        \\ 
    \rowcolor{gray!5}  
    DeCoF\textsuperscript{\textdagger\textdagger}   & 91.69             & 96.30              & 97.18              & 76.02              & 70.40             & 59.39              & 93.77                 & 91.18               & 98.68                & 98.70                  & 98.43                 & 48.99               & 86.75                & 83.85                  & 85.10               \\
    NSG-VD\textsuperscript{\textdagger}             & 65.31 & 49.64 & 50.32 & 62.49 & 57.42 & 55.39 & 70.45 & 59.14 & 51.59 & 51.86 & 52.56 & 56.84 & 68.55 & 67.21 & 58.48  \\
    \rowcolor{gray!5}
    ReStraV\textsuperscript{\textdagger}            & 97.72             & \textbf{99.93}              & \underline{99.94}              & \textbf{99.21}              & \underline{95.70}             & \underline{97.35}             & 70.64             & 96.04             & 96.99             & \textbf{100.00}             & 99.71       & 77.14             & 75.18             & 63.45             & 90.64              \\ 
    D3\textsuperscript{\textdagger}                 & 94.09           & 93.36           & 96.22           & 96.96           & 89.37           & 87.28           & 96.17           & 86.70           & 96.74           & 94.34           & 95.02           & 94.16           & 95.78           & 94.84           & 93.64       \\ 
    \rowcolor{gray!5}  
    STALL\textsuperscript{\textdagger}              & 83.60            & 88.52            & 88.72            & 77.79            & 86.39            & 82.84            & 85.35            & 77.75            & 75.54            & 82.85            & 88.20            & 80.20            & 78.02            & 73.36            & 82.08            \\
    \hline
    STPAG\textsuperscript{\textdagger\textdagger}   & 81.54 & 89.98 & 91.27 & 86.70 & 82.21 & 82.94 & 92.81 & 78.20 & 93.87 & 94.88 & 94.19 & 93.53 & 86.37 & 68.72 & 86.94 \\ 
    \rowcolor{gray!5} 
    LGN-Net\textsuperscript{\textdagger}            & 87.81 & 93.22 & 98.62 & 97.06 & 87.41 & 87.43 & 95.25 & 89.10 & 96.41 & 96.67 & 93.82 & 96.29 & 92.43 & 89.58 & 92.94 \\  
    X-CLIP\textsuperscript{\textdagger}             & 87.32        & 87.56        & 90.72        & 89.67        & 85.58        & 83.55        & 94.22        & 79.45        & 96.04        & 97.44        & 96.15        & 95.68        & 91.42        & 77.63        & 90.28        \\     
    \hline \hline
    \rowcolor{gray!25}
    \textbf{MTOR}    & \underline{99.33}       & \underline{99.70}       & \textbf{99.95}       & \underline{99.12}       & \textbf{99.10}       & \textbf{97.63}       & \textbf{99.93}       & \textbf{98.66}       & \underline{99.91}       & 99.91       & \underline{99.83}       & \textbf{98.87}       & \underline{99.52}       & \textbf{98.27}       & \textbf{99.27}       \\
    \bottomrule
    \end{tabular}
    }
\label{tab:ap_EvalCrafter}
\end{table*}

\begin{table*}[th]
    \centering
    \caption{Comparison of AUC (\%) between MTOR and 16 representative baselines on EvalCrafter.  \textsuperscript{\textdagger} Results reproduced using the official code. \textsuperscript{\textdagger\textdagger} Results reproduced from our implementation because no official code is available.}
    \resizebox{\textwidth}{!}{
    \begin{tabular}{c | c c c c c c c c c c c c c c| c}
    \toprule
    \rowcolor{gray!25}
    Method &  \makecell[c]{\rotatebox{50}{Floor33}} & \makecell[c]{\rotatebox{50}{Gen2}} &  \makecell[c]{\rotatebox{50}{Gen2-D}} &  \makecell[c]{\rotatebox{50}{HS-XL}} &  \makecell[c]{\rotatebox{50}{LaVie-B}} &  \makecell[c]{\rotatebox{50}{LaVie-I}} &  \makecell[c]{\rotatebox{50}{Mix-SR}} &  \makecell[c]{\rotatebox{50}{MS}} &  \makecell[c]{\rotatebox{50}{MV}} &  \makecell[c]{\rotatebox{50}{PKL}} &  \makecell[c]{\rotatebox{50}{PKL-V1}} &  \makecell[c]{\rotatebox{50}{Show-1}} &  \makecell[c]{\rotatebox{50}{VC}} & \makecell[c]{\rotatebox{50}{ZS}} &  \makecell[c]{\rotatebox{50}{mean}} \\
    \midrule
    \hline
    \rowcolor{gray!5} 
    FID\textsuperscript{\textdagger}                & 96.26 & 97.46 & 98.58 & 90.05 & 92.32     & 83.42     & 98.23 & 95.27     & 98.28 & 99.48     & 99.24 & \underline{96.79} & 95.26 & \underline{95.29}     & \underline{95.42}     \\ 
    NPR\textsuperscript{\textdagger}                & \textbf{99.38}     & 97.74             & 99.80     & 26.83             & 50.23             & 36.00             & \underline{99.22}             & 93.92             & \textbf{99.92}     & \underline{99.92}     & 99.71     & 18.99             & \underline{99.25}     & 92.60     & 79.54             \\ 
    \hline
    \rowcolor{gray!5}
    STIL\textsuperscript{\textdagger}               & 90.14             & 91.40             & 79.25             & 61.62             & 73.63             & 71.23             & 74.22             & 87.28             & 84.22             & 97.82     & 88.50             & 66.68             & 74.49             & 90.77     & 80.80          \\  
    FTCN\textsuperscript{\textdagger}               & 82.78             & 90.76             & 95.11             & 90.29             & 84.51             & 83.00             & 92.31             & 70.99             & 96.65             & 96.73     & 94.01             & 93.62             & 91.27             & 70.82      & 88.06          \\
    \rowcolor{gray!5}
    TALL\textsuperscript{\textdagger}               & 61.46             & 72.04             & 74.60             & 43.88             & 49.64             & 50.02             & 77.50             & 58.99             & 83.19             & 67.13             & 72.51             & 46.81             & 55.69             & 46.03             & 61.39             \\ 
    MINTIME\textsuperscript{\textdagger}           & 86.88             & 88.01             & 89.41             & 91.71             & 84.07             & 84.23             & 90.16             & 79.60             & 91.68             & 95.08      & 92.33            & 90.08               & 90.16            & 91.29      & 88.91               \\
    \hline
    \rowcolor{gray!5}
    AIGVDet\textsuperscript{\textdagger}            & 79.52             & 70.10             & 73.03             & 73.92             & 85.34 & 74.73 & 64.44             & 68.17             & 59.90             & 92.78             & 90.60             & 70.23             & 74.92             & 67.75             & 74.67             \\ 
    DeMamba\textsuperscript{\textdagger}                      & 95.41             & 98.60     & 99.70     & 59.10             & 72.87             & 55.39             & 98.91     & 79.11             & 99.87     & 99.51 & 99.70     & 48.89             & 97.08             & 76.98             & 84.37             \\
    \rowcolor{gray!5}  
    DeCoF\textsuperscript{\textdagger\textdagger}   & 96.87             & 98.69              & 99.17              & 90.22              & 87.71             & 81.53              & 98.05                 & 95.49               & 99.60                & 99.54                  & 99.42                 & 77.89               & 95.20                & 92.20                  & 93.68               \\
    NSG-VD\textsuperscript{\textdagger}             & 74.71 & 59.07 & 61.21 & 74.14 & 67.80 & 66.42 & 79.27 & 69.46 & 59.74 & 61.52 & 63.72 & 69.77 & 76.69 & 77.39 & 68.64  \\
    \rowcolor{gray!5}
    ReStraV\textsuperscript{\textdagger}            & 98.33             & \textbf{99.93}              & \underline{99.94}              & \underline{99.13}              & \underline{96.93}             & \textbf{97.97}              & 73.98              & \underline{97.43}              & 96.53             & \textbf{100.00}              & \textbf{99.85}              & 81.71              & 82.05             & 68.22             & 92.29              \\   
    D3\textsuperscript{\textdagger}                 & 93.68           & 92.24           & 95.85           & 96.93           & 90.06           & 88.68           & 95.62           & 87.38           & 96.32           & 93.77           & 94.73           & 92.49           & 95.58           & 93.76           & 93.36              \\ 
    \rowcolor{gray!5}  
    STALL\textsuperscript{\textdagger}              & 82.60            & 86.62            & 88.21            & 78.79            & 86.02            & 84.00            & 83.70            & 77.64            & 72.03            & 82.14            & 86.81            & 81.79            & 80.05            & 72.46            & 81.63            \\  
    \hline
    STPAG\textsuperscript{\textdagger\textdagger}   & 82.76 & 91.43 & 93.86 & 88.65 & 83.18 & 83.55 & 93.34 & 79.72 & 94.91 & 95.77 & 94.77 & 94.47 & 87.59 & 74.07 & 88.43 \\ 
    \rowcolor{gray!5} 
    LGN-Net\textsuperscript{\textdagger}            & 87.56 & 95.32 & 98.76 & 97.35 & 87.85 & 87.98 & 95.43 & 88.79 & 96.74 & 97.90 & 95.07 & 96.53 & 92.68 & 88.25 & 93.30 \\ 
    X-CLIP\textsuperscript{\textdagger}             & 86.10             & 86.70             & 90.93             & 90.80             & 84.06             & 83.01             & 93.59             & 79.69             & 95.56             & 97.45     & 95.79             & 95.29             & 90.71             & 80.53      & 89.38          \\  
    \hline \hline
    \rowcolor{gray!25}
    \textbf{MTOR}    & \underline{99.31}       & \underline{99.74}       & \textbf{99.95}       & \textbf{99.17}       & \textbf{99.04}       & \underline{97.67}       & \textbf{99.93}       & \textbf{98.67}       & \underline{99.91}       & 99.91       & \underline{99.84}       & \textbf{98.89}       & \textbf{99.54}       & \textbf{98.56}       & \textbf{99.30}       \\
    \bottomrule
    \end{tabular}
    }
\label{tab:auc_EvalCrafter}
\end{table*}

\begin{table*}[th]
    \centering
    \caption{Comparison of AP (\%) between MTOR and 16 representative baselines on VideoPhy.  \textsuperscript{\textdagger} Results reproduced using the official code. \textsuperscript{\textdagger\textdagger} Results reproduced from our implementation because no official code is available.}
    \resizebox{\textwidth}{!}{
    \begin{tabular}{c | c c c c c c c c c c| c}
    \toprule
    \rowcolor{gray!25}
    Method                                                                                                 & CVX   & CVX-5B   & DM   & Gen-2        & LaVie        & OpenSora     & Pika       & SVD-T2I2V    & VC2     & ZS    & mean   \\
    \midrule
    \hline
    \rowcolor{gray!5}
    FID\textsuperscript{\textdagger} \cite{NEURIPS2024_6dddcff5}, NeurIPS 2024                             & \underline{93.34}       & 91.41         & 97.50   & 98.35 & \underline{96.51} & 87.90 & 99.55 & 95.66 & 96.03     & 90.60 & \underline{94.69} \\
    NPR\textsuperscript{\textdagger} \cite{Tan_2024_CVPR}, CVPR 2024                                       & 81.37        & 81.99         & \textbf{99.86}   & 99.90 & 63.72        & 88.78 & 99.91 & \underline{99.54}        & 60.21             & 78.23        & 85.35   \\
    \hline
    \rowcolor{gray!5}
    STIL\textsuperscript{\textdagger} \cite{10.1145/3474085.3475508}, ACM MM 2021                          & 66.43        & 69.61         & 65.31           & 69.63        & 63.68        & 61.94        & 97.07      & 67.51        & 67.54             & 58.56        & 68.73   \\
    FTCN\textsuperscript{\textdagger} \cite{zheng2021exploring}, ICCV 2021                                 & 74.24        & 74.83         & 67.98           & 93.99        & 69.16        & 66.95        & 93.94      & 82.81        & 83.39             & 75.64        & 78.29   \\
    \rowcolor{gray!5} 
    TALL\textsuperscript{\textdagger} \cite{xu2023tall}, ICCV 2023                                         & 39.59        & 50.72         & 62.36           & 70.78        & 40.40        & 37.30        & 62.69      & 52.62        & 52.66             & 50.66        & 51.98   \\
    MINTIME\textsuperscript{\textdagger} \cite{10547206}, TIFS 2024                                       & 85.27             & 85.96              & 77.60              & 84.12              & 79.16             & 90.64              & 92.86                 & 79.01               & 83.83                & 86.34                  & 84.48               \\ 
    \hline
    \rowcolor{gray!5}
    AIGVDet\textsuperscript{\textdagger} \cite{bai2024ai}, PRCV 2024                                       & 63.15        & 58.95         & 59.27           & 61.55        & 61.06        & 59.07        & 92.96      & 53.73        & 58.22             & 63.11        & 63.11   \\
    DeMamba\textsuperscript{\textdagger} \cite{chen2024demamba}, arXiv 2024                                & 22.10        & 15.96         & 72.52           & 92.92        & 50.00        & 94.70        & 45.47      & 91.59        & 90.15             & 73.64        & 64.91   \\
    \rowcolor{gray!5}
    DeCoF\textsuperscript{\textdagger\textdagger} \cite{Ma2024DetectingAV}, ICME 2025                      & 14.35             & 17.10              & 79.20              & 88.81              & 42.72             & 19.92              & 95.01                 & 59.68               & 46.13                & 21.59                  & 48.45               \\
    NSG-VD\textsuperscript{\textdagger}~\cite{zhang2025NSGVD}, NeurIPS 2025               & 53.77 & 56.28 & 55.59 & 49.43 & 58.79 & 60.17 & 50.04 & 63.52 & 58.63 & 61.58 & 56.78  \\
    \rowcolor{gray!5}
    ReStraV\textsuperscript{\textdagger} \cite{interno2025aigenerated}, NeurIPS 2025                       & 72.57             & 61.39              & 99.09              & \textbf{99.95}              & \textbf{98.61}             & \textbf{99.57}             & \textbf{100.00}             & 45.01             & \underline{99.42}             & \textbf{98.55}             & 87.42              \\ 
    D3\textsuperscript{\textdagger} \cite{Zheng_2025_ICCV}, ICCV 2025                                      & 90.23        & \underline{93.94}         & 96.25           & 94.88           & 82.22           & 92.38           & 93.21           & 95.60           & 92.50           & 92.52           & 92.37              \\ 
    \rowcolor{gray!5}  
    STALL\textsuperscript{\textdagger}  \cite{hayun2026stall}, CVPR 2026                                   & 66.13            & 64.49            & 80.50            & 85.55            & 80.84            & 74.70            & 80.91            & 74.96            & 82.87            & 74.40            & 76.54            \\  
    \hline
    STPAG\textsuperscript{\textdagger\textdagger}  \cite{Rai_2024_CVPR}, CVPRW 2024   & 78.08 & 75.36 & 67.41 & 87.23 & 69.89 & 74.25 & 94.36 & 69.37 & 76.60 & 69.60 & 76.21 \\ 
    \rowcolor{gray!5} 
    LGN-Net\textsuperscript{\textdagger}  \cite{ZHAO2025125581}, ESWA 2025            & 86.28 & 84.67 & 75.06 & 95.20 & 80.59 & 88.81 & 95.94 & 89.70 & 87.69 & 93.46 & 87.74 \\
    X-CLIP\textsuperscript{\textdagger} \cite{ni2022expanding}, ECCV 2022                                  & 87.35        & 83.72         & 75.54           & 92.09        & 79.52        & 90.44        & 96.13      & 85.26        & 85.35             & 88.88        & 86.66   \\     
    \hline \hline
    \rowcolor{gray!25}
    \textbf{MTOR}    & \textbf{96.10}       & \textbf{96.88}       & \underline{99.50}       & \underline{99.94}       & 96.18       & \underline{97.19}       & \underline{99.99}       & \textbf{99.83}       & \textbf{99.50}       & \underline{96.36}       & \textbf{98.15}       \\
    \bottomrule
    \end{tabular}
    }
\label{tab:ap_VideoPhy}
\end{table*}

\begin{table*}[th]
    \centering
    \caption{Comparison of AUC (\%) between MTOR and 16 representative baselines on VideoPhy.  \textsuperscript{\textdagger} Results reproduced using the official code. \textsuperscript{\textdagger\textdagger} Results reproduced from our implementation because no official code is available.}
    \resizebox{\textwidth}{!}{
    \begin{tabular}{c | c c c c c c c c c c | c}
    \toprule
    \rowcolor{gray!25}
    Method                                                                                                 & CVX    & CVX-5B & DM   & Gen-2   & LaVie   & OpenSora      & Pika    & SVD-T2I2V   & VC2   & ZS    & mean   \\
    \midrule
    \hline
    \rowcolor{gray!5}
    FID\textsuperscript{\textdagger} \cite{NEURIPS2024_6dddcff5}, NeurIPS 2024                             & \underline{93.37} & 91.72        & 97.62   & 98.11 & \underline{96.03} & 86.03 & 99.59 & 95.81 & 95.68     & 89.38 & \underline{94.33} \\
    NPR\textsuperscript{\textdagger} \cite{Tan_2024_CVPR}, CVPR 2024                                       & 72.10        & 73.60        & \textbf{99.70}   & \underline{99.80} & 42.90   & 83.50        & 99.80 & \underline{99.50}       & 47.20           & 52.90        & 77.10        \\
    \hline
    \rowcolor{gray!5}
    STIL\textsuperscript{\textdagger} \cite{10.1145/3474085.3475508}, ACM MM 2021                          & 71.73        & 73.94        & 73.01           & 80.75   & 68.66   & 63.42        & 98.47 & 77.82       & 75.18           & 55.85        & 73.88        \\
    FTCN\textsuperscript{\textdagger} \cite{zheng2021exploring}, ICCV 2021                                 & 80.91        & 81.01        & 70.20           & 95.21   & 72.88   & 76.37        & 95.90 & 87.07       & 86.53           & 74.88        & 82.10        \\
    \rowcolor{gray!5}
    TALL\textsuperscript{\textdagger} \cite{xu2023tall}, ICCV 2023                                         & 30.05        & 46.36        & 56.54           & 64.80   & 32.94   & 24.82        & 63.65    & 45.40       & 42.48           & 39.44        & 44.65        \\
    MINTIME\textsuperscript{\textdagger} \cite{10547206}, TIFS 2024                                       & 85.08        & 87.24        & 82.06           & 86.32          & 76.29   & 89.92        & 93.61          & 80.36                & 85.09           & 89.30        & 85.53        \\
    \hline
    \rowcolor{gray!5}
    AIGVDet\textsuperscript{\textdagger} \cite{bai2024ai}, PRCV 2024                                       & 78.13        & 67.05        & 68.40           & 75.87   & 71.25   & 69.57        & 97.83    & 72.87       & 81.36           & 77.15        & 75.95        \\
    DeMamba\textsuperscript{\textdagger} \cite{chen2024demamba}, arXiv 2024                                          & 59.65        & 61.44        & 98.42           & 99.64   & 63.66   & 51.32        & 99.19    & 97.72 & 63.57           & 46.56        & 74.12        \\
    \rowcolor{gray!5}
    DeCoF\textsuperscript{\textdagger\textdagger} \cite{Ma2024DetectingAV}, ICME 2025                      & 53.29             & 60.92              & 96.21              & 98.64              & 82.43             & 64.90              & 99.10                 & 92.99               & 84.98                & 63.71                  & 79.72               \\
    NSG-VD\textsuperscript{\textdagger}~\cite{zhang2025NSGVD}, NeurIPS 2025            & 64.24 & 68.41 & 68.96 & 58.81 & 70.63 & 72.21 & 60.11 & 75.83 & 72.06 & 72.76 & 68.40  \\
    \rowcolor{gray!5}
    ReStraV\textsuperscript{\textdagger} \cite{interno2025aigenerated}, NeurIPS 2025                       & 78.09             & 65.21              & 98.83              & \textbf{99.94}              & \textbf{98.91}             & \textbf{99.65}             & \textbf{100.00}             & 41.02             & \underline{99.44}             & \textbf{99.01}             & 88.01              \\
    D3\textsuperscript{\textdagger} \cite{Zheng_2025_ICCV}, ICCV 2025                                      & 90.05           & \underline{93.51}           & 95.56           & 94.34           & 84.26           & 91.90           & 92.02           & 95.09           & 91.68           & 90.91           & 91.93            \\
    \rowcolor{gray!5}
    STALL\textsuperscript{\textdagger}  \cite{hayun2026stall}, CVPR 2026                                   & 66.11            & 64.10            & 77.64            & 84.77            & 78.09            & 70.61            & 76.71            & 75.31            & 79.88            & 72.96            & 74.62            \\  
    \hline
    STPAG\textsuperscript{\textdagger\textdagger}  \cite{Rai_2024_CVPR}, CVPRW 2024   & 79.84 & 79.42 & 72.07 & 90.78 & 68.23 & 79.84 & 95.05 & 79.86 & 82.10 & 77.30 & 80.45 \\ 
    \rowcolor{gray!5} 
    LGN-Net\textsuperscript{\textdagger}  \cite{ZHAO2025125581}, ESWA 2025            & 89.49 & 87.85 & 78.15 & 96.84 & 80.05 & 91.72 & 96.48 & 93.15 & 90.05 & 93.58 & 89.74 \\ 
    X-CLIP\textsuperscript{\textdagger} \cite{ni2022expanding}, ECCV 2022                                  & 88.99        & 84.84        & 78.57           & 93.35   & 76.51   & 92.68        & 96.22 & 84.92       & 86.27           & 88.06        & 86.88        \\     
    \hline \hline
    \rowcolor{gray!25}
    \textbf{MTOR}                                                                                         & \textbf{96.22}       & \textbf{97.46}       & \underline{99.58}       & \textbf{99.94}       & 95.67       & \underline{97.49}       & \underline{99.99}       & \textbf{99.83}       & \textbf{99.56}       & \underline{97.61}       & \textbf{98.33}       \\
    \bottomrule
    \end{tabular}
    }
\label{tab:auc_VideoPhy}
\end{table*}

\subsection{Baselines}
To comprehensively evaluate the performance of the proposed MTOR, we compare it against 16 representative baselines, consisting of 2 image-level detectors and 14 video-level frameworks. 
According to their target domains and foundational paradigms, the video-level detectors are rigorously partitioned into three primary categories: 
(i) deepfake video detection methods, including STIL, FTCN, TALL, and MINTIME; 
(ii) AI-generated video detection frameworks, which include learning-based networks such as AIGVDet, DeMamba, DeCoF, and NSG-VD, alongside training-free detection models such as ReStraV, D3, and STALL; and 
(iii) cross-domain adapted approaches, including STPAG and LGN-Net from the video anomaly detection domain, as well as X-CLIP from the general video recognition domain, all adjusted with a binary head for synthetic video classification. 
A concise conceptual overview of each baseline is detailed below:
\begin{itemize}

    \item \textbf{FID}~\cite{NEURIPS2024_6dddcff5}: Leverages a patch-shuffle mechanism to eliminate semantic artifacts for AI-generated image detection, extending to video detection by temporal average pooling.
    
    \item \textbf{NPR}~\cite{Tan_2024_CVPR}: Targets CNN-specific up-sampling artifacts to expose AI-generated images, with video-level predictions aggregated via temporal pooling strategy.
    
    \item \textbf{STIL}~\cite{10.1145/3474085.3475508}: Capitalizes on a spatiotemporal inconsistency learning block to capture multi-directional temporal dynamics and spatial artifacts across frames.
    
    \item \textbf{FTCN}~\cite{zheng2021exploring}: Utilizes $1\times1$ spatial convolutions and a Transformer to capture both short- and long-term temporal coherence for general face forgery detection.
    
    \item \textbf{TALL}~\cite{xu2023tall}: Rearranges consecutive frames into a pre-defined thumbnail layout to efficiently capture spatiotemporal dependencies using model-agnostic backbones.
    
    \item \textbf{MINTIME}~\cite{10547206}: Performs multi-identity deepfake detection by capturing spatiotemporal inconsistencies across identities and encoding face-scale contextual cues.

    \item \textbf{AIGVDet}~\cite{bai2024ai}: Detects AI-generated videos via a two-branch framework capturing spatial anomalies in RGB frames and temporal patterns in optical flow.
    
    \item \textbf{DeMamba}~\cite{chen2024demamba}: Leverages a Mamba-based network to efficiently process long-range, frame-wise temporal dependencies for synthetic video detection.
    
    \item \textbf{DeCoF}~\cite{11210049}: Identifies AI-generated videos by modeling inter-frame consistency in the visual feature space to capture temporal artifacts.

    \item \textbf{NSG-VD}~\cite{zhang2025NSGVD}: Exposes AI-generated videos by computing a Normalized Spatiotemporal Gradient based on probability flow conservation, and leveraging Maximum Mean Discrepancy to detect physical dynamics deviations.

    \item \textbf{ReStraV}~\cite{interno2025aigenerated}: Uncovers AI-generated videos via perceptual straightening in representation space, using temporal curvature and stepwise distance.

    \item \textbf{D3}~\cite{Zheng_2025_ICCV}: Discriminates AI-generated videos via a training-free framework by computing second-order central difference features to capture temporal artifacts.

    \item \textbf{STALL}~\cite{hayun2026stall}: Introduces a training-free paradigm that distinguishes AI-generated videos from real ones by combining separate spatial and temporal likelihood scores into a joint probabilistic framework.

    \item \textbf{STPAG}~\cite{Rai_2024_CVPR}: Detects AI-generated videos by adapting an anomaly detection framework that captures generative artifacts via diffusion models and optical perturbations.

    \item \textbf{LGN-Net}~\cite{ZHAO2025125581}: Exposes generative artifacts by adapting a memory-augmented, local-global spatiotemporal anomaly detection framework for AI-generated video detection.
    
    \item \textbf{X-CLIP}~\cite{ni2022expanding}: Employs a frozen, cross-frame attention-based spatiotemporal backbone coupled with a binary classifier to distinguish synthetic videos.
\end{itemize}

To ensure a rigorous and fair comparison, all baselines and our MTOR are evaluated under identical dataset splits and unified training and testing protocols.

\begin{table*}[th]
    \centering
    \caption{Comparison of AP (\%) between MTOR and 16 representative baselines on VidProM. \textsuperscript{\textdagger} Results reproduced using the official code. \textsuperscript{\textdagger\textdagger} Results reproduced from our implementation because no official code is available.}
    \resizebox{0.70\textwidth}{!}{
    \begin{tabular}{c | c c c c c c | c}
    \toprule
    \rowcolor{gray!25}
    Method                                                                                    & MS          & OpenSora    & Pika        & ST2V        &  T2VZ       & VC2         & mean \\
    \midrule
    \hline
    \rowcolor{gray!5}
    FID\textsuperscript{\textdagger} \cite{NEURIPS2024_6dddcff5}, NeurIPS 2024                & 91.35       & 87.68       & 99.59       & \textbf{97.87} & 68.51    & 85.92       & 88.49       \\
    NPR\textsuperscript{\textdagger} \cite{Tan_2024_CVPR}, CVPR 2024                          & 87.04       & 89.85       & \textbf{99.98} & 89.88    & 88.93       & 70.79       & 87.75       \\
    \hline
    \rowcolor{gray!5}
    STIL\textsuperscript{\textdagger} \cite{10.1145/3474085.3475508}, ACM MM 2021             & 43.68       & 63.15       & 95.28       & 55.46       & 48.68       & 61.49       & 61.29       \\
    FTCN\textsuperscript{\textdagger} \cite{zheng2021exploring}, ICCV 2021                    & 74.65       & 90.05       & 94.26       & 83.53       & 46.25       & 87.27       & 79.33       \\
    \rowcolor{gray!5}
    TALL\textsuperscript{\textdagger} \cite{xu2023tall}, ICCV 2023                            & 50.93       & 54.50       & 63.47       & 51.50       & 60.99       & 59.70       & 56.85       \\
    MINTIME\textsuperscript{\textdagger} \cite{10547206}, TIFS 2024                           & 76.09       & 88.32       & 94.19       & 50.03       & 65.06       & 86.77       & 76.74       \\
    \hline
    \rowcolor{gray!5}
    AIGVDet\textsuperscript{\textdagger} \cite{bai2024ai}, PRCV 2024                          & 63.33       & 62.12       & 66.07       & 55.46       & 63.49       & 52.15       & 60.44       \\
    DeMamba\textsuperscript{\textdagger} \cite{chen2024demamba}, arXiv 2024                   & \textbf{99.24}   & 48.00  & 86.95  & 33.94  & \textbf{98.47}  & \underline{98.38}  & 77.50   \\
    \rowcolor{gray!5}
    DeCoF\textsuperscript{\textdagger\textdagger} \cite{Ma2024DetectingAV}, ICME 2025         & 87.93       & 85.42       & 99.56       & 76.27       & 93.72  & 96.75  & \underline{90.11}     \\
    NSG-VD\textsuperscript{\textdagger}~\cite{zhang2025NSGVD}, NeurIPS 2025                                & 65.69 & 55.29 & 51.55 & 58.52 & \underline{96.44} & 53.83 & 63.55  \\
    \rowcolor{gray!5}
    ReStraV\textsuperscript{\textdagger} \cite{interno2025aigenerated}, NeurIPS 2025          & \underline{96.99} & \underline{97.20}   & 99.85    & 63.96  & 39.63 & 69.14  & 77.80            \\ 
    D3\textsuperscript{\textdagger} \cite{Zheng_2025_ICCV}, ICCV 2025                         & 87.61       & 90.73       & 93.17       & 81.00       & 61.93       & 93.97       & 84.73       \\  
    \rowcolor{gray!5} 
    STALL\textsuperscript{\textdagger}  \cite{hayun2026stall}, CVPR 2026                      & 87.04       & 81.46       & 82.92       & 77.45       & 70.73       & 83.06       & 80.44       \\ 
    \hline
    STPAG\textsuperscript{\textdagger\textdagger}  \cite{Rai_2024_CVPR}, CVPRW 2024           & 62.77       & 91.26       & 92.36       & 80.01       & 52.20       & 76.87        & 75.91       \\ 
    \rowcolor{gray!5} 
    LGN-Net\textsuperscript{\textdagger}  \cite{ZHAO2025125581}, ESWA 2025                    & 81.79       & 93.05       & 95.45       & 73.70       & 57.09       & 90.79        & 81.98 \\ 
    X-CLIP\textsuperscript{\textdagger} \cite{ni2022expanding}, ECCV 2022                     & 82.24       & 93.47       & 95.80       & 86.42       & 70.92       & 91.60       & 88.05       \\   
    \hline \hline
    \rowcolor{gray!25}
    \textbf{MTOR}                                                                            & 94.12       & \textbf{97.56}       & \underline{99.91}       & \underline{92.24}       & 95.13       & \textbf{99.51}       & \textbf{96.41}        \\
    \bottomrule
    \end{tabular}
    }
\label{tab:ap_VidProM}
\end{table*}

\begin{table*}[th]
    \centering
    \caption{Comparison of AUC (\%) between MTOR and 16 representative baselines on VidProM. \textsuperscript{\textdagger} Results reproduced using the official code. \textsuperscript{\textdagger\textdagger} Results reproduced from our implementation because no official code is available.}
    \resizebox{0.70\textwidth}{!}{
    \begin{tabular}{c | c c c c c c | c}
    \toprule
    \rowcolor{gray!25}
    Method                                                                                                 & MS           & OpenSora   & Pika        & ST2V     &  T2VZ    & VC2     & mean \\
    \midrule
    \hline
    \rowcolor{gray!5}
    FID\textsuperscript{\textdagger} \cite{NEURIPS2024_6dddcff5}, NeurIPS 2024                             & 89.86        & 87.36       & 99.62 & \underline{98.10} & 67.01       & 85.65 & 87.93 \\
    NPR\textsuperscript{\textdagger} \cite{Tan_2024_CVPR}, CVPR 2024                                       & 82.61        & \textbf{98.56} & \underline{99.84} & \textbf{98.92} & 93.32 & 56.70       & 88.33 \\
    \hline
    \rowcolor{gray!5}
    STIL\textsuperscript{\textdagger} \cite{10.1145/3474085.3475508}, ACM MM 2021                          & 41.05        & 72.01       & 97.42 & 67.12       & 47.66       & 67.98       & 65.54       \\
    FTCN\textsuperscript{\textdagger} \cite{zheng2021exploring}, ICCV 2021                                 & 70.98        & 90.76       & 95.23       & 87.81       & 50.70       & 88.43       & 80.65       \\
    \rowcolor{gray!5}
    TALL\textsuperscript{\textdagger} \cite{xu2023tall}, ICCV 2023                                         & 45.29        & 56.39       & 66.91       & 50.31       & 57.80       & 53.48       & 55.03       \\
    MINTIME\textsuperscript{\textdagger} \cite{10547206}, TIFS 2024                                        & 79.40        & 88.99       & 94.81       & 54.21       & 70.92       & 88.96       & 79.55       \\
    \hline
    \rowcolor{gray!5}
    AIGVDet\textsuperscript{\textdagger} \cite{bai2024ai}, PRCV 2024                                       & 60.11        & 45.27       & 48.75       & 34.14       & 59.96       & 46.21       & 49.07       \\
    DeMamba\textsuperscript{\textdagger} \cite{chen2024demamba}, arXiv 2024                                & 54.44        & 84.02       & 99.29       & 84.94       & 76.34       & 78.61       & 79.61       \\
    \rowcolor{gray!5}
    DeCoF\textsuperscript{\textdagger\textdagger} \cite{Ma2024DetectingAV}, ICME 2025                      & 89.18        & 86.98       & 99.56       & 80.14       & 93.85       & \underline{96.94}       & \underline{90.77}       \\
    NSG-VD\textsuperscript{\textdagger}~\cite{zhang2025NSGVD}, NeurIPS 2025                                & 75.19        & 65.02       & 61.34       & 66.54       & \textbf{97.08}       & 61.06       & 71.04  \\
    \rowcolor{gray!5}
    ReStraV\textsuperscript{\textdagger} \cite{interno2025aigenerated}, NeurIPS 2025                       & \textbf{97.96}             & \underline{97.53}             & \underline{99.87}             & 68.44             & 28.60             & 73.79             & 77.70              \\ 
    D3\textsuperscript{\textdagger} \cite{Zheng_2025_ICCV}, ICCV 2025                                      & 87.21        & 91.29       & 92.24       & 79.74       & 56.31       & 92.77       & 83.26         \\  
    \rowcolor{gray!5}   
    STALL\textsuperscript{\textdagger}  \cite{hayun2026stall}, CVPR 2026                                   & 85.32        & 81.75            & 81.41            & 76.66            & 70.98            & 80.37            & 79.42            \\  
    \hline
    STPAG\textsuperscript{\textdagger\textdagger}  \cite{Rai_2024_CVPR}, CVPRW 2024                        & 66.73        & 90.91       & 93.37       & 86.13       & 57.08       & 80.35      & 79.10         \\ 
    \rowcolor{gray!5} 
    LGN-Net\textsuperscript{\textdagger}  \cite{ZHAO2025125581}, ESWA 2025                                 & 78.77        & 93.43       & 96.14       & 79.34       & 63.27       & 90.45       & 83.57 \\ 
    X-CLIP\textsuperscript{\textdagger} \cite{ni2022expanding}, ECCV 2022                                  & 80.59        & 93.02       & 95.85       & 87.47       & 75.86       & 91.24       & 87.34       \\ 
    \hline \hline
    \rowcolor{gray!25}
    \textbf{MTOR}                                                                                         & \underline{94.85}       & 97.46       & \textbf{99.92}       & 92.57       & \underline{95.34}       & \textbf{99.53}       & \textbf{96.61}       \\ 
    \bottomrule
    \end{tabular}
    }
\label{tab:auc_VidProM}
\end{table*}

\subsection{Evaluation Metrics}
Following prior state-of-the-art methods, such as ReStraV~\cite{interno2025aigenerated}, NSG-VD~\cite{zhang2025NSGVD}, DeMamba~\cite{chen2024demamba}, and DeCoF~\cite{Ma2024DetectingAV}, we adopt three widely used metrics to evaluate our approach: Average Precision (AP), Area Under the Receiver Operating Characteristic curve (AUC), and Accuracy (ACC). AP evaluates the precision-recall trade-off via the area under the precision-recall curve, reflecting the model's ability to precisely rank synthetic videos. AUC quantifies the overall detection capability across all possible thresholds, representing the probability that a randomly chosen AI-generated video is scored higher than a real one. ACC measures the proportion of correctly classified real and AI-generated videos.
Throughout Tables~\ref{tab:ap_GenVideo}--\ref{tab:ablation2_acc}, bold and underlined values denote the best and second-best results within each column, respectively.

\subsection{Quantitative Results}
We comprehensively evaluate the proposed MTOR against 16 state-of-the-art baselines across 5 large-scale video benchmarks, including GenVideo, EvalCrafter, VideoPhy, VidProM, and GenVidBench. The quantitative results in terms of AP, AUC, and ACC are summarized in Tables~\ref{tab:ap_GenVideo}--\ref{tab:acc_all}. 
For baselines with officially released code, we reproduce their results using the official implementations. For methods without official implementations, we carefully re-implement them following the experimental settings in the original papers.

\begin{table*}[th]
    \centering
    \caption{Comparison of AP (\%) between MTOR and 16 representative baselines on GenVidBench.  \textsuperscript{\textdagger} Results reproduced using the official code. \textsuperscript{\textdagger\textdagger} Results reproduced from our implementation because no official code is available.}
    \resizebox{0.7\textwidth}{!}{
    \begin{tabular}{c || c c c c c c| c}
    \toprule
    \rowcolor{gray!25}
    Method                                                                                     & CogVideo    & SVD         & Sora        &  Mora       & MuseV       & Kling       & mean \\
    \midrule
    \hline
    \rowcolor{gray!5}
    FID\textsuperscript{\textdagger} \cite{NEURIPS2024_6dddcff5}, NeurIPS 2024                 & 73.10       & 55.95       & 80.06       & 49.13       & 57.29       & 77.29       & 65.47       \\
    NPR\textsuperscript{\textdagger} \cite{Tan_2024_CVPR}, CVPR 2024                           & 77.88       & 72.85       & 82.56       & 82.13       & 80.27       & 90.58       & 81.05       \\
    \hline
    \rowcolor{gray!5}
    STIL\textsuperscript{\textdagger} \cite{10.1145/3474085.3475508}, ACM MM 2021              & 72.83       & 63.00       & 52.00       & 63.84       & 64.44       & 57.90       & 62.34       \\
    FTCN\textsuperscript{\textdagger} \cite{zheng2021exploring}, ICCV 2021                     & 85.96       & 87.17       & 44.40       & 74.11       & 97.57       & 74.50       & 77.29       \\
    \rowcolor{gray!5}
    TALL\textsuperscript{\textdagger} \cite{xu2023tall}, ICCV 2023                             & 55.08       & 59.38       & 51.64       & 55.66       & 58.78       & 57.61       & 56.36       \\
    MINTIME\textsuperscript{\textdagger} \cite{10547206}, TIFS 2024                            & 85.70       & 75.72       & 76.62       & 69.53       & 91.05       & 88.57       & 81.20       \\ 
    \hline
    \rowcolor{gray!5}
    AIGVDet\textsuperscript{\textdagger} \cite{bai2024ai}, PRCV 2024                           & \textbf{97.79}       & 76.36       & 12.71       & 92.59       & 89.68       & 40.08       & 68.20       \\
    DeMamba\textsuperscript{\textdagger} \cite{chen2024demamba}, arXiv 2024                    & 78.00       & 68.87       & 88.68       & \underline{97.73}       & 65.19       & 72.26       & 78.46       \\
    \rowcolor{gray!5}
    DeCoF\textsuperscript{\textdagger\textdagger} \cite{Ma2024DetectingAV}, ICME 2025          & 87.71       & 69.93       & 91.56       & 96.50       & 78.56       & 89.26       & 85.59       \\
    NSG-VD\textsuperscript{\textdagger} \cite{zhang2025NSGVD}, NeurIPS 2025                    & 64.76       & 61.60       & 53.78       & 63.01       & 52.28       & 53.44       & 58.15       \\
    \rowcolor{gray!5}
    ReStraV\textsuperscript{\textdagger} \cite{interno2025aigenerated}, NeurIPS 2025           & 70.20       & 62.58       & \textbf{98.97}       & 61.81       & 95.17       & \textbf{99.26}       & 81.33       \\ 
    D3\textsuperscript{\textdagger} \cite{Zheng_2025_ICCV}, ICCV 2025                          & \underline{94.28}       & 91.57       & 83.84       & 93.88       & 89.33       & 93.40       & \underline{91.05}       \\  
    \rowcolor{gray!5}
    STALL\textsuperscript{\textdagger}  \cite{hayun2026stall}, CVPR 2026                       & 75.80       & 57.48       & 73.66       & 85.33       & 65.02       & 78.64       & 72.66       \\ 
    \hline
    STPAG\textsuperscript{\textdagger\textdagger}  \cite{Rai_2024_CVPR}, CVPRW 2024            & 85.97       & 91.25       & 50.80       & 75.52       & 94.85       & 79.68       & 79.68       \\ 
    \rowcolor{gray!5} 
    LGN-Net\textsuperscript{\textdagger}  \cite{ZHAO2025125581}, ESWA 2025                     & 92.25       & \textbf{96.26}       & 55.16       & 93.92       & \textbf{98.10}       & 86.91       & 87.10       \\ 
    X-CLIP\textsuperscript{\textdagger} \cite{ni2022expanding}, ECCV 2022                      & 88.25       & 91.58       & 63.93       & 77.27       & 96.72       & 83.67       & 83.57       \\ 
    \hline \hline 
    \rowcolor{gray!25}
    \textbf{MTOR}                                                                             & 93.06       & \underline{95.06}       & \underline{96.75}       & \textbf{98.51}       & \underline{97.91}       & \underline{96.18}       & \textbf{96.24}       \\ 
    \bottomrule
    \end{tabular}
    }
\label{tab:ap_GenVidBench}
\end{table*}

\begin{table*}[th]
    \centering
    \caption{Comparison of AUC (\%) between MTOR and 16 representative baselines on GenVidBench. \textsuperscript{\textdagger} Results reproduced using the official code. \textsuperscript{\textdagger\textdagger} Results reproduced from our implementation because no official code is available.}
    \resizebox{0.7\textwidth}{!}{
    \begin{tabular}{c || c c c c c c| c}
    \toprule
    \rowcolor{gray!25}
    Method                                                                                     & CogVideo    & SVD         & Sora        &  Mora       & MuseV       & Kling       & mean \\
    \midrule
    \hline
    \rowcolor{gray!5}
    FID\textsuperscript{\textdagger} \cite{NEURIPS2024_6dddcff5}, NeurIPS 2024                 & 76.05       & 53.99       & 77.32       & 40.49       & 54.65       & 76.84       & 63.22       \\
    NPR\textsuperscript{\textdagger} \cite{Tan_2024_CVPR}, CVPR 2024                           & 74.85       & 65.67       & 77.82       & 75.27       & 73.14       & 88.74       & 75.92       \\
    \hline
    \rowcolor{gray!5}
    STIL\textsuperscript{\textdagger} \cite{10.1145/3474085.3475508}, ACM MM 2021              & 74.90       & 57.78       & 57.09       & 61.02       & 60.22       & 65.07       & 62.68       \\
    FTCN\textsuperscript{\textdagger} \cite{zheng2021exploring}, ICCV 2021                     & 90.15       & 90.87       & 38.91       & 77.56       & \underline{97.98}       & 77.41       & 78.81       \\
    \rowcolor{gray!5}
    TALL\textsuperscript{\textdagger} \cite{xu2023tall}, ICCV 2023                             & 44.12       & 50.88       & 56.63       & 42.29       & 50.09       & 55.51       & 49.92       \\
    MINTIME\textsuperscript{\textdagger} \cite{10547206}, TIFS 2024                            & 85.69       & 78.83       & 76.32       & 74.86       & 91.35       & 89.46       & 82.75       \\ 
    \hline
    \rowcolor{gray!5}
    AIGVDet\textsuperscript{\textdagger} \cite{bai2024ai}, PRCV 2024                           & \textbf{96.81}       & 69.85       & 80.80       & 87.77       & 84.87       & 86.02       & 84.35       \\
    DeMamba\textsuperscript{\textdagger} \cite{chen2024demamba}, arXiv 2024                    & 77.58       & 64.87       & 87.27       & \underline{97.16}       & 62.69       & 63.77       & 75.56       \\
    \rowcolor{gray!5}
    DeCoF\textsuperscript{\textdagger\textdagger} \cite{Ma2024DetectingAV}, ICME 2025          & 87.45       & 70.95       & 91.67       & 96.57       & 78.09       & 88.45       & 85.53       \\
    NSG-VD\textsuperscript{\textdagger} \cite{zhang2025NSGVD}, NeurIPS 2025                    & 76.16       & 71.62       & 71.79       & 72.50       & 62.13       & 66.48       & 70.11       \\
    \rowcolor{gray!5}
    ReStraV\textsuperscript{\textdagger} \cite{interno2025aigenerated}, NeurIPS 2025           & 70.01       & 65.31       & \textbf{98.85}       & 67.04       & 94.95       & \textbf{99.26}       & 82.57       \\ 
    D3\textsuperscript{\textdagger} \cite{Zheng_2025_ICCV}, ICCV 2025                          & \underline{94.08}       & 92.68       & 86.97       & 95.13       & 90.05       & 94.54       & \underline{92.24}       \\ 
    \rowcolor{gray!5}
    STALL\textsuperscript{\textdagger}  \cite{hayun2026stall}, CVPR 2026                       & 73.97       & 58.92       & 72.05       & 84.20       & 66.65       & 79.11       & 72.48       \\ 
    \hline
    STPAG\textsuperscript{\textdagger\textdagger}  \cite{Rai_2024_CVPR}, CVPRW 2024            & 83.32       & 91.06       & 49.40       & 80.21       & 94.47       & 81.71       & 80.03       \\ 
    \rowcolor{gray!5} 
    LGN-Net\textsuperscript{\textdagger}  \cite{ZHAO2025125581}, ESWA 2025                     & 91.46       & \textbf{96.71}       & 53.52       & 94.04       & 97.95       & 87.06       & 86.79       \\  
    X-CLIP\textsuperscript{\textdagger} \cite{ni2022expanding}, ECCV 2022                      & 89.76       & 92.47       & 62.78       & 77.33       & 96.91       & 82.59       & 83.64       \\  
    \hline \hline 
    \rowcolor{gray!25}
    \textbf{MTOR}                                                                             & 93.90       & \underline{95.93}       & \underline{96.58}       & \textbf{98.58}       & \textbf{98.20}       & \underline{97.05}       & \textbf{96.71}       \\ 
    \bottomrule
    \end{tabular}
    }
\label{tab:auc_GenVidBench}
\end{table*}

\begin{table}[!ht]
    \centering
    \caption{Comparison of ACC (\%) between MTOR and 16 representative baselines across five benchmarks.
    \textsuperscript{\textdagger} Results reproduced using the official code. \textsuperscript{\textdagger\textdagger} Results reproduced from our implementation because no official code is available.}
    \resizebox{\columnwidth}{!}{
    \begin{tabular}{c | c c c c c}
    \toprule
    \rowcolor{gray!25}
    Method                                                      & GenVideo           & EvalCrafter        & VideoPhy            & VidProM            & GenVidBench\\
    \midrule
    \hline
    \rowcolor{gray!5}
    FID\textsuperscript{\textdagger}                            & 54.57              & 63.59              & 65.01               & 54.44              & 57.24  \\  
    NPR\textsuperscript{\textdagger}                            & 65.41              & 71.36              & 57.00               & 68.04              & 68.39  \\
    \hline
    \rowcolor{gray!5}
    STIL\textsuperscript{\textdagger}                           & 59.90              & 66.78              & 56.62               & 56.58              & 46.09  \\
    FTCN\textsuperscript{\textdagger}                           & 70.52              & 73.87              & 62.43               & 67.23              & 64.88  \\
    \rowcolor{gray!5}  
    TALL\textsuperscript{\textdagger}                           & 57.47              & 58.42              & 48.78               & 54.02              & 49.23  \\
    MINTIME\textsuperscript{\textdagger}                        & 78.55              & 81.52              & 77.13               & 71.74              & 73.74  \\
    \hline
    \rowcolor{gray!5}
    AIGVDet\textsuperscript{\textdagger}                        & 49.07              & 57.62              & 53.33               & 47.25              & 76.02  \\ 
    DeMamba\textsuperscript{\textdagger}                        & 54.12              & 62.45              & 42.29               & 42.59              & 66.73  \\  
    \rowcolor{gray!5}
    DeCoF\textsuperscript{\textdagger\textdagger}               & \underline{87.60}  & \underline{89.75}  & 63.71               & \underline{85.47}  & 70.41  \\ 
    NSG-VD\textsuperscript{\textdagger}                         & 58.17              & 57.88              & 57.83               & 58.20              & 57.23  \\ 
    \rowcolor{gray!5}
    ReStraV\textsuperscript{\textdagger}                        & 56.17              & 63.93              & 62.00               & 57.13              & 51.30  \\
    D3\textsuperscript{\textdagger}                             & 76.19              & 76.88              & 72.84               & 64.16              & \underline{82.89}  \\ 
    \rowcolor{gray!5}  
    STALL\textsuperscript{\textdagger}                          & 56.19              & 56.73              & 56.44               & 57.04              & 55.61  \\  
    \hline
    STPAG\textsuperscript{\textdagger\textdagger}               & 75.65              & 78.38              & 65.18               & 68.26              & 66.59  \\ 
    \rowcolor{gray!5} 
    LGN-Net\textsuperscript{\textdagger}                        & 82.59              & 86.09              & \underline{78.07}   & 73.29              & 79.79  \\ 
    X-CLIP\textsuperscript{\textdagger}                         & 75.34              & 76.82              & 69.26               & 73.37              & 69.86  \\       
    \hline \hline
    \rowcolor{gray!25}
    \textbf{MTOR}                                              & \textbf{93.36}     & \textbf{95.48}     & \textbf{90.70}      & \textbf{86.48}     & \textbf{85.18}  \\
    \bottomrule
    \end{tabular}
    }
\label{tab:acc_all}
\end{table}

\begin{table}[!t]
    \centering
    \caption{Ablation results of MTOR across five benchmarks in terms of AP (\%).}
    \resizebox{\columnwidth}{!}{
    \begin{tabular}{c | c c c c c | c}
    \toprule
    Variant
    & \rotatebox[origin=c]{45}{GenVideo}
    & \rotatebox[origin=c]{45}{EvalCrafter}
    & \rotatebox[origin=c]{45}{VideoPhy}
    & \rotatebox[origin=c]{45}{VidProM}
    & \rotatebox[origin=c]{45}{GenVidBench}
    & Mean \\
    \midrule

    Visual-only
    & 97.42
    & 98.06
    & 91.59
    & \underline{96.58}
    & 90.24
    & 94.78 \\

    Text-only
    & 93.18
    & 95.08
    & 90.76
    & 89.76
    & 88.60
    & 91.47 \\

    Multimodal $G$
    & \textbf{98.81}
    & \underline{99.11}
    & \underline{96.61}
    & \textbf{97.11}
    & \underline{94.66}
    & \underline{97.26} \\

    TOR-only
    & 94.28
    & 96.28
    & 95.13
    & 89.28
    & 91.14
    & 93.22 \\

    \textbf{MTOR}
    & \underline{98.74}
    & \textbf{99.27}
    & \textbf{98.15}
    & 96.41
    & \textbf{96.24}
    & \textbf{97.76} \\

    \bottomrule
    \end{tabular}
    }
    \label{tab:ablation_ap}
\end{table}

\begin{table}[!ht]
    \centering
    \caption{Ablation results of MTOR across five benchmarks in terms of AUC (\%).}
    \resizebox{\columnwidth}{!}{
    \begin{tabular}{c | c c c c c | c}
    \toprule
    Variant
    & \rotatebox[origin=c]{45}{GenVideo}
    & \rotatebox[origin=c]{45}{EvalCrafter}
    & \rotatebox[origin=c]{45}{VideoPhy}
    & \rotatebox[origin=c]{45}{VidProM}
    & \rotatebox[origin=c]{45}{GenVidBench}
    & Mean \\
    \midrule

    Visual-only
    & 97.04
    & 97.83
    & 90.75
    & 96.27
    & 89.36
    & 94.25 \\

    Text-only
    & 92.81
    & 95.34
    & 92.67
    & 88.69
    & 89.08
    & 91.72 \\

    Multimodal $G$
    & \textbf{98.66}
    & \underline{99.08}
    & \underline{96.50}
    & \textbf{96.91}
    & \underline{94.69}
    & \underline{97.17} \\

    TOR-only
    & 95.02
    & 97.12
    & 95.98
    & 91.65
    & 93.18
    & 94.59 \\

    \textbf{MTOR}
    & \underline{98.64}
    & \textbf{99.30}
    & \textbf{98.33}
    & \underline{96.61}
    & \textbf{96.71}
    & \textbf{97.92} \\

    \bottomrule
    \end{tabular}
    }
    \label{tab:ablation_auc}
\end{table}

\begin{table}[!ht]
    \centering
    \caption{Ablation results of MTOR across five benchmarks in terms of ACC (\%).}
    \resizebox{\columnwidth}{!}{
    \begin{tabular}{c | c c c c c | c}
    \toprule
    Variant
    & \rotatebox[origin=c]{45}{GenVideo}
    & \rotatebox[origin=c]{45}{EvalCrafter}
    & \rotatebox[origin=c]{45}{VideoPhy}
    & \rotatebox[origin=c]{45}{VidProM}
    & \rotatebox[origin=c]{45}{GenVidBench}
    & Mean \\
    \midrule

    Visual-only
    & 90.30
    & 91.03
    & 83.58
    & \textbf{89.40}
    & \underline{82.42}
    & 87.35 \\

    Text-only
    & 85.64
    & 88.51
    & 81.66
    & 82.17
    & 78.64
    & 83.32 \\

    Multimodal $G$
    & \textbf{93.80}
    & \underline{95.08}
    & \underline{87.87}
    & \underline{88.92}
    & 81.86
    & \underline{89.51} \\

    TOR-only
    & 86.14
    & 90.03
    & 86.07
    & 78.59
    & 80.14
    & 84.19 \\

    \textbf{MTOR}
    & \underline{93.36}
    & \textbf{95.48}
    & \textbf{90.70}
    & 86.48
    & \textbf{85.18}
    & \textbf{90.24} \\

    \bottomrule
    \end{tabular}
    }
    \label{tab:ablation_acc}
\end{table}

\subsubsection{GenVideo}
Tables~\ref{tab:ap_GenVideo} and~\ref{tab:auc_GenVideo} compare MTOR with 16 baseline methods on GenVideo in terms of AP and AUC, respectively, while Table~\ref{tab:acc_all} reports the corresponding ACC results.
MTOR achieves the highest mean AP, AUC, and ACC of 98.74\%, 98.64\%, and 93.36\%, respectively.
Specifically, MTOR surpasses the strongest baseline D3 by 6.58 percentage points in mean AP.
In terms of mean AUC, our approach exceeds the second-best competitor ReStraV by 6.15 percentage points.
Furthermore, MTOR outperforms the closest competitor DeCoF by 5.76 percentage points in ACC.

The baseline methods exhibit substantial performance variations across different generator subsets.
Deepfake detectors, originally developed for facial manipulation, generally show limited transferability to open-domain AI-generated videos on GenVideo.
Image-level detectors remain competitive on certain generators, but their performance varies considerably across subsets.
For instance, NPR achieves only 21.61\% AP and 18.99\% AUC on the Show-1 subset, and 22.37\% AP and 42.32\% AUC on the LaVie subset.

At the generator level, MTOR achieves the best AP on 7 of the 10 subsets and maintains consistently high performance across different video generators.
Overall, these results demonstrate the strong cross-generator generalization of MTOR across diverse video synthesis models.

\subsubsection{EvalCrafter}
We evaluate the comprehensive performance on 14 generator subsets of EvalCrafter, where Table~\ref{tab:ap_EvalCrafter}, Table~\ref{tab:auc_EvalCrafter} and Table~\ref{tab:acc_all} present the AP, AUC and ACC results, respectively.
MTOR achieves the highest mean AP and AUC of 99.27\% and 99.30\%, outperforming the strongest baseline FID by 3.68 and 3.88 percentage points, respectively.
It also achieves the highest ACC of 95.48\%, exceeding the second-best baseline, DeCoF, by 5.73 percentage points.

EvalCrafter is built from a diverse prompt set covering different semantic categories, styles, and motion settings, providing varied generated content for evaluation.
Across the 14 generator subsets, MTOR maintains AP and AUC above 97.6\% on every subset.
It achieves the highest AP on 7 subsets, including a tie on Gen2-D, and the highest AUC on 8 subsets.
Its lowest performance occurs on LaVie-Interpolation, where it still achieves 97.63\% AP and 97.67\% AUC.
Overall, these results demonstrate the consistently strong performance of MTOR across diverse generator subsets and generated content.

\subsubsection{VideoPhy}
To assess detection robustness under diverse physical-interaction scenarios, we conduct evaluations on the 10 subsets of VideoPhy, reporting the AP, AUC, and ACC metrics in Table~\ref{tab:ap_VideoPhy}, Table~\ref{tab:auc_VideoPhy}, and Table~\ref{tab:acc_all}, respectively.
MTOR achieves the highest mean AP and AUC of 98.15\% and 98.33\%, outperforming the second-best baseline FID by 3.46 and 4.00 percentage points, respectively.
It also achieves the highest ACC of 90.70\%, exceeding the second-ranked baseline LGN-Net by 12.63 percentage points.

Among the baselines, FID achieves the highest mean AP and AUC, indicating that image-level spatial cues can remain effective on VideoPhy.
This is consistent with the visible artifacts associated with some physical failures in VideoPhy, such as inconsistent geometry and texture and unnatural object deformation.
At the same time, many physical violations involve object motion and interactions that evolve over time, making temporal behavior an important aspect of this benchmark.
MTOR maintains AP above 96.1\% and AUC above 95.6\% across all 10 generator subsets, demonstrating consistently strong performance across diverse video generators and physical-interaction scenarios.

\subsubsection{VidProM}
To evaluate detection performance on videos generated from diverse real-user prompts, we conduct experiments on the VidProM dataset.
The corresponding AP, AUC, and ACC metrics are reported in Table~\ref{tab:ap_VidProM}, Table~\ref{tab:auc_VidProM}, and Table~\ref{tab:acc_all}, respectively.
MTOR achieves the highest mean AP and AUC of 96.41\% and 96.61\%, outperforming the second-best baseline DeCoF by 6.30 and 5.84 percentage points, respectively.
It also achieves the highest ACC of 86.48\%, exceeding DeCoF by 1.01 percentage points.
Moreover, MTOR maintains strong performance across all six generator subsets, demonstrating robust cross-generator generalization under diverse video-generation conditions.

VidProM is constructed from real-user prompts with broad semantic diversity, which often describe temporal changes and dynamic object behaviors.
This provides a semantically diverse and dynamic evaluation setting, under which the baseline methods exhibit considerable performance variations across generator subsets.
For example, DeCoF, despite achieving the strongest mean AP and AUC among the baselines, ranges from 76.27\% to 99.56\% in AP and from 80.14\% to 99.56\% in AUC.
In comparison, MTOR maintains AP above 92.2\% and AUC above 92.5\% across all six subsets, resulting in substantially higher mean performance.
Overall, these results demonstrate the consistent performance of MTOR across diverse generator subsets with real-user prompts.

\subsubsection{GenVidBench}
To assess cross-generator generalization, we conduct extensive evaluations on six generator subsets from GenVidBench.
Tables~\ref{tab:ap_GenVidBench} and \ref{tab:auc_GenVidBench} report the AP and AUC scores across the six subsets, while Table~\ref{tab:acc_all} summarizes the overall ACC results.
MTOR achieves the highest mean AP and AUC of 96.24\% and 96.71\%, outperforming the second-best baseline D3 by 5.19 and 4.47 percentage points, respectively.
It also achieves the highest ACC of 85.18\%, exceeding D3 by 2.29 percentage points.
Moreover, MTOR maintains AP and AUC above 93\% across all six generator subsets, demonstrating strong robustness under substantial generator variations.

The six subsets span diverse generation settings, including text-to-video and image-to-video generation with different resolutions and frame rates.
In particular, MuseV and SVD are image-to-video models, Mora, CogVideo, and Sora are text-to-video models, while Kling supports both settings.
These heterogeneous generation settings provide a challenging test of cross-generator generalization.
While several baselines achieve strong performance on individual generators, their results vary considerably across subsets.
In contrast, MTOR maintains a consistently high performance level across all six generators, resulting in the highest overall mean AP and AUC.

\begin{table*}[!t]
    \centering
    \caption{Ablation results on GenVideo for different captioning models and visual-textual encoders in terms of AP (\%).}
    \resizebox{\textwidth}{!}{
    \begin{tabular}{c c | c c c c c c c c c c | c}
    \toprule
    Caption Model   & Visual \& Textual Encoder    & MS       & MPS       & MV       & HotShot       & Show-1       & Gen2       & Crafter       & LaVie       & Sora       & WS       & mean   \\
    \midrule 
    \hline
    \multirow{4}{*}{BLIP-base}
              & CLIP-P16                       & 90.00 & 94.26 & 99.48 & 93.52 & 95.33 & 99.45 & 98.44 & 92.55 & 78.17 & 89.75 & 93.09 \\
              & CLIP-P32                       & 84.53 & 91.26 & 98.84 & 96.12 & 91.73 & 99.21 & 97.64 & 90.03 & 83.61 & 89.01 & 92.20 \\
              & XCLIP-P16                      & 91.12 & 95.32 & 99.40 & 96.96 & 96.29 & 99.32 & 98.58 & 94.54 & 79.92 & 91.49 & 94.29 \\
              & XCLIP-P32                      & 82.03 & 90.66 & 98.57 & 94.15 & 93.87 & 98.43 & 97.10 & 91.46 & 72.30 & 87.08 & 90.57 \\
    \hline
    \multirow{4}{*}{BLIP-large}
              & CLIP-P16                       & 88.18 & 93.16 & 99.31 & 92.11 & 94.15 & 99.28 & 98.01 & 91.52 & 75.76 & 88.50 & 92.00 \\
              & CLIP-P32                       & 85.63 & 92.35 & 98.80 & 95.14 & 94.52 & 99.22 & 97.65 & 91.21 & 71.29 & 90.56 & 91.64 \\
              & XCLIP-P16                      & 89.50 & 94.02 & 99.24 & 96.35 & 95.41 & 99.20 & 98.14 & 93.91 & 80.40 & 90.71 & 93.69 \\
              & XCLIP-P32                      & 81.58 & 90.15 & 98.36 & 93.81 & 93.32 & 98.24 & 96.78 & 91.35 & 72.88 & 86.86 & 90.33 \\
    \hline
    \multirow{2}{*}{CogVLM2-Caption}  
              & LanguageBind                   & 96.52    & 97.12     & 99.49    & 98.56         & 97.91        & 99.67      & 99.09         & 97.73       & 88.92      & 96.08    & 97.11   \\
              & InternVideo 2.5                & \underline{98.86}    & \textbf{99.33}     & \textbf{99.96}    & \textbf{99.17}         & \underline{98.53}        & \underline{99.91}      & \textbf{99.76}         & 98.06       & \underline{96.63}      & \underline{97.02}    & \underline{98.72}   \\
    \hline
    \multirow{2}{*}{UGC-VideoCaptioner}  
              & LanguageBind                   & 96.35    & 97.40     & 99.44    & 97.16         & 97.34        & 99.79      & 98.95         & 97.27       & 93.82      & 94.24    & 97.18   \\
              & InternVideo 2.5                & 98.32    & 98.93     & 99.88    & \underline{98.70}         & 98.01        & \textbf{99.92}      & 99.68         & \underline{98.09}       & \textbf{97.91}      & 96.84    & 98.63   \\
    \hline 
    \multirow{2}{*}{ShareGPT4Video-8B}    
              & LanguageBind                   & 92.81    & 94.80     & 98.72    & 95.33         & 95.07        & 98.37      & 97.71         & 95.43       & 85.48      & 93.23    & 94.70   \\ 
              & \cellcolor{gray!25}InternVideo 2.5        & \cellcolor{gray!25}\textbf{98.91}  & \cellcolor{gray!25}\underline{99.22}  & \cellcolor{gray!25}\underline{99.89}   & \cellcolor{gray!25}\textbf{99.17}  & \cellcolor{gray!25}\textbf{98.79}    & \cellcolor{gray!25}99.84    & \cellcolor{gray!25}\underline{99.72}     & \cellcolor{gray!25}\textbf{98.26}      & \cellcolor{gray!25}96.59    & \cellcolor{gray!25}\textbf{97.06}          & \cellcolor{gray!25}\textbf{98.74}  \\ 
    \bottomrule
    \end{tabular}
    }
\label{tab:ablation2_ap}
\end{table*}

\begin{table*}[!t]
    \centering
    \caption{Ablation results on GenVideo for different captioning models and visual-textual encoders in terms of AUC (\%).}
    \resizebox{\textwidth}{!}{
    \begin{tabular}{c c | c c c c c c c c c c | c}
    \toprule
    Caption Model   & Visual \& Textual Encoder    & MS       & MPS       & MV       & HotShot       & Show-1       & Gen2       & Crafter       & LaVie       & Sora       & WS       & mean   \\
    \midrule    
    \hline
    \multirow{4}{*}{BLIP-base}
              & CLIP-P16                       & 90.47 & 93.55 & 99.42 & 93.91 & 95.10 & 99.43 & 98.21 & 91.57 & 78.09 & 88.64 & 92.84 \\
              & CLIP-P32                       & 86.25 & 90.66 & 98.70 & 96.21 & 92.85 & 99.18 & 97.40 & 89.71 & 84.12 & 88.91 & 92.40 \\
              & XCLIP-P16                      & 91.64 & 94.97 & 99.33 & 97.09 & 96.42 & 99.30 & 98.39 & 93.85 & 81.82 & 91.06 & 94.39 \\
              & XCLIP-P32                      & 84.26 & 90.45 & 98.39 & 94.88 & 94.16 & 98.29 & 96.76 & 90.89 & 73.82 & 87.19 & 90.91 \\
    \hline
    \multirow{4}{*}{BLIP-large}
              & CLIP-P16                       & 89.00 & 92.35 & 99.25 & 92.88 & 93.99 & 99.27 & 97.71 & 90.63 & 75.99 & 87.62 & 91.87 \\
              & CLIP-P32                       & 87.04 & 91.71 & 98.67 & 95.29 & 94.60 & 99.18 & 97.41 & 90.53 & 72.32 & 90.39 & 91.71 \\
              & XCLIP-P16                      & 90.47 & 93.65 & 99.17 & 96.55 & 95.54 & 99.17 & 97.90 & 93.24 & 80.64 & 90.28 & 93.66 \\
              & XCLIP-P32                      & 84.48 & 90.06 & 98.25 & 94.69 & 93.77 & 98.18 & 96.52 & 91.07 & 73.28 & 87.38 & 90.77 \\
    \hline
    \multirow{2}{*}{CogVLM2-Caption}  
              & LanguageBind                   & 96.57    & 96.85     & 99.53    & 98.67         & 97.87        & 99.72      & 99.01         & 97.66       & 87.53      & 95.35    & 96.88   \\
              & InternVideo 2.5                & \underline{98.83}    & \textbf{99.32}     & \textbf{99.96}    & \underline{99.16}         & \underline{98.62}        & \textbf{99.92}      & \textbf{99.77}         & 97.92       & 95.25      & \underline{96.80}    & \underline{98.56}   \\
    \hline
    \multirow{2}{*}{UGC-VideoCaptioner}  
              & LanguageBind                   & 96.29    & 97.15     & 99.44    & 97.97         & 97.72        & 99.80      & 98.90         & 97.45       & 92.19      & 93.43    & 97.03   \\
              & InternVideo 2.5                & 98.33    & 98.97     & 99.87    & 98.64         & 98.35        & \textbf{99.92}      & 99.68         & \underline{98.02}       & \textbf{97.23}      & 96.46    & 98.55   \\
    \hline 
    \multirow{2}{*}{ShareGPT4Video-8B}    
              & LanguageBind                   & 93.63    & 94.74     & 98.83    & 96.68         & 96.06        & 98.80      & 97.84         & 95.93       & 84.44      & 92.60    & 94.95   \\ 
              & \cellcolor{gray!25}InternVideo 2.5        & \cellcolor{gray!25}\textbf{98.90}  & \cellcolor{gray!25}\underline{99.24}  & \cellcolor{gray!25}\underline{99.89}   & \cellcolor{gray!25}\textbf{99.21}  & \cellcolor{gray!25}\textbf{98.80}    & \cellcolor{gray!25}\underline{99.85}    & \cellcolor{gray!25}\textbf{99.74}     & \cellcolor{gray!25}\textbf{98.22}      & \cellcolor{gray!25}\underline{95.60}    & \cellcolor{gray!25}\textbf{96.92}          & \cellcolor{gray!25}\textbf{98.64}  \\ 
    \bottomrule
    \end{tabular}
    }
\label{tab:ablation2_auc}
\end{table*}

\begin{table*}[!t]
    \centering
    \caption{Ablation results on GenVideo for different captioning models and visual-textual encoders in terms of ACC (\%).}
    \resizebox{\textwidth}{!}{
    \begin{tabular}{c c | c c c c c c c c c c | c}
    \toprule
    Caption Model   & Visual \& Textual Encoder    & MS       & MPS       & MV       & HotShot       & Show-1       & Gen2       & Crafter       & LaVie       & Sora       & WS       & mean   \\
    \midrule    
    \hline
    \multirow{4}{*}{BLIP-base}
              & CLIP-P16                       & 75.14 & 83.50 & 96.01 & 80.29 & 85.86 & 96.05 & 93.53 & 80.79 & 62.50 & 76.55 & 83.02 \\
              & CLIP-P32                       & 67.07 & 78.50 & 93.85 & 86.36 & 80.36 & 95.80 & 90.77 & 77.00 & 66.96 & 73.21 & 80.99 \\
              & XCLIP-P16                      & 76.00 & 85.29 & 95.85 & 88.00 & 88.21 & 96.16 & 93.28 & 84.71 & 71.43 & 77.81 & 85.67 \\
              & XCLIP-P32                      & 66.50 & 77.93 & 92.89 & 84.50 & 84.21 & 92.83 & 89.91 & 80.93 & 59.82 & 72.96 & 80.25 \\
    \hline
    \multirow{4}{*}{BLIP-large}
              & CLIP-P16                       & 75.21 & 82.86 & 94.57 & 80.29 & 84.86 & 95.14 & 92.49 & 80.68 & 63.39 & 76.10 & 82.56 \\
              & CLIP-P32                       & 71.71 & 81.36 & 93.29 & 86.43 & 86.57 & 94.64 & 91.17 & 80.50 & 60.71 & 78.75 & 82.51 \\
              & XCLIP-P16                      & 74.86 & 83.29 & 94.81 & 88.29 & 87.50 & 95.36 & 92.63 & 83.96 & 66.96 & 76.82 & 84.45 \\
              & XCLIP-P32                      & 67.36 & 78.29 & 92.49 & 85.79 & 84.43 & 92.75 & 89.48 & 82.00 & 58.04 & 74.50 & 80.51 \\
    \hline
    \multirow{2}{*}{CogVLM2-Caption}  
              & LanguageBind                   & 82.57    & 87.50     & 97.04    & 90.14         & 89.57        & 98.15      & 95.06         & 88.96       & 69.64      & 82.01    & 88.07   \\
              & InternVideo 2.5                & \underline{91.36}    & \textbf{95.50}     & \textbf{98.72}    & \textbf{94.86}         & \underline{91.00}        & \underline{98.41}      & \textbf{97.85}         & \underline{90.07}       & \textbf{89.29}      & \underline{85.90}    & \underline{93.30}   \\
    \hline
    \multirow{2}{*}{UGC-VideoCaptioner}  
              & LanguageBind                   & 79.71    & 85.21     & 95.77    & 86.71         & 86.36        & 97.50      & 94.03         & 86.14       & 67.86      & 77.57    & 85.69   \\
              & InternVideo 2.5                & 82.93    & 91.00     & 98.32    & 89.21         & 83.79        & \textbf{98.62}      & 96.10         & 85.21       & 83.04      & 77.26    & 88.55   \\
    \hline 
    \multirow{2}{*}{ShareGPT4Video-8B}    
              & LanguageBind                   & 79.21    & 85.21     &  94.33    & 89.57         & 88.43        & 95.62      & 92.70         & 87.75       & 69.64      & 79.52    & 86.20   \\ 
              & \cellcolor{gray!25}InternVideo 2.5        & \cellcolor{gray!25}\textbf{92.07}  & \cellcolor{gray!25}\underline{95.14}  & \cellcolor{gray!25}\underline{98.48}   & \cellcolor{gray!25}\underline{94.71}  & \cellcolor{gray!25}\textbf{93.21}    & \cellcolor{gray!25}97.68    & \cellcolor{gray!25}\underline{97.28}     & \cellcolor{gray!25}\underline{91.39}      & \cellcolor{gray!25}\underline{87.50}    & \cellcolor{gray!25}\textbf{86.14}          & \cellcolor{gray!25}\textbf{93.36}  \\ 
    \bottomrule
    \end{tabular}
    }
\label{tab:ablation2_acc}
\end{table*}

\subsection{Ablation Studies}
We conduct two sets of ablation studies to examine the key design choices of MTOR.
1) We evaluate the contributions of visual and textual representations, their multimodal combination, and the TOR component across all five benchmarks, with the results reported in Tables~\ref{tab:ablation_ap}--\ref{tab:ablation_acc}.
2) We investigate different combinations of captioning models and visual-textual encoders on GenVideo, as reported in Tables~\ref{tab:ablation2_ap}--\ref{tab:ablation2_acc}.

\subsubsection{Contributions of the Multimodal Branch and TOR Component}
Tables~\ref{tab:ablation_ap}--\ref{tab:ablation_acc} compare five variants: visual-only, text-only, the multimodal branch $G$, the TOR-only variant, and the complete MTOR.
The visual-only variant achieves five-benchmark mean AP, AUC, and ACC of 94.78\%, 94.25\%, and 87.35\%, respectively, outperforming the text-only variant in overall mean performance.
Combining the global visual and caption-derived textual representations in the multimodal branch $G$ substantially improves the mean performance to 97.26\% AP, 97.17\% AUC, and 89.51\% ACC.
Compared with the visual-only variant, these correspond to gains of 2.48, 2.92, and 2.16 percentage points.
Notably, the multimodal combination improves AP and AUC over the visual-only variant on all five benchmarks, demonstrating the complementary value of caption-derived textual semantics for video-level detection.

When used independently, the TOR-only variant achieves mean AP, AUC, and ACC of 93.22\%, 94.59\%, and 84.19\%, respectively.
Its complementary contribution becomes more evident when combined with the multimodal branch.
Compared with the individual multimodal branch $G$, the complete MTOR improves the mean AP, AUC, and ACC across the five benchmarks by 0.50, 0.75, and 0.73 percentage points, respectively.
The largest gains are observed on VideoPhy and GenVidBench, where MTOR improves AP by 1.54 and 1.58 percentage points, AUC by 1.83 and 2.02 percentage points, and ACC by 2.83 and 3.32 percentage points, respectively.
Although performance decreases occur on some benchmarks, most notably on VidProM, the overall mean performance improves across all three metrics.
These results indicate that the TOR component provides complementary temporal cues beyond those captured by the multimodal representations.

\begin{figure*}[ht]
    \centering
    \captionsetup[subfloat]{font=footnotesize, labelfont=small}  
    \subfloat[Crafter]{\includegraphics[width=0.19\textwidth]{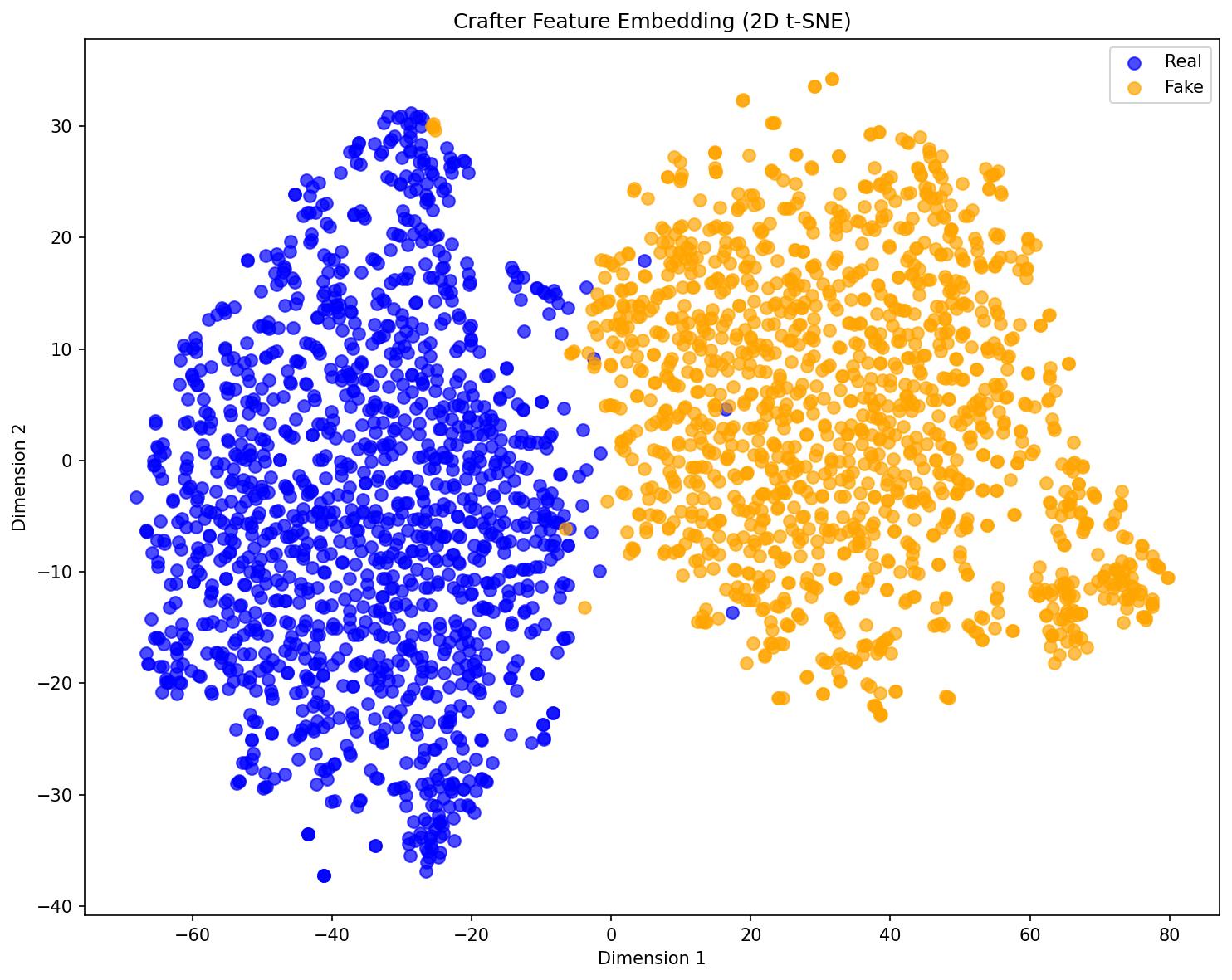}}\hfill
    \subfloat[Gen2]{\includegraphics[width=0.19\textwidth]{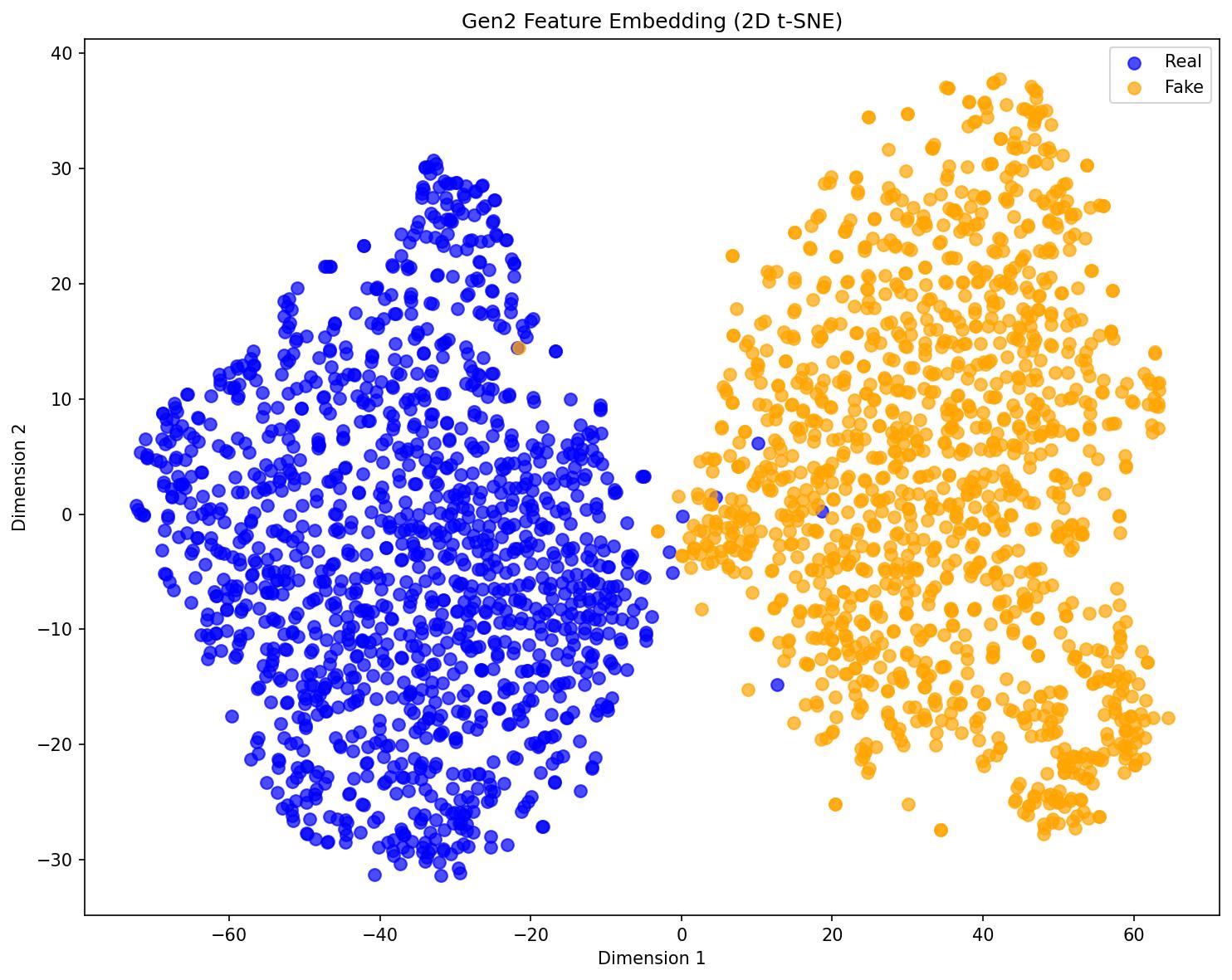}}\hfill
    \subfloat[HotShot]{\includegraphics[width=0.19\textwidth]{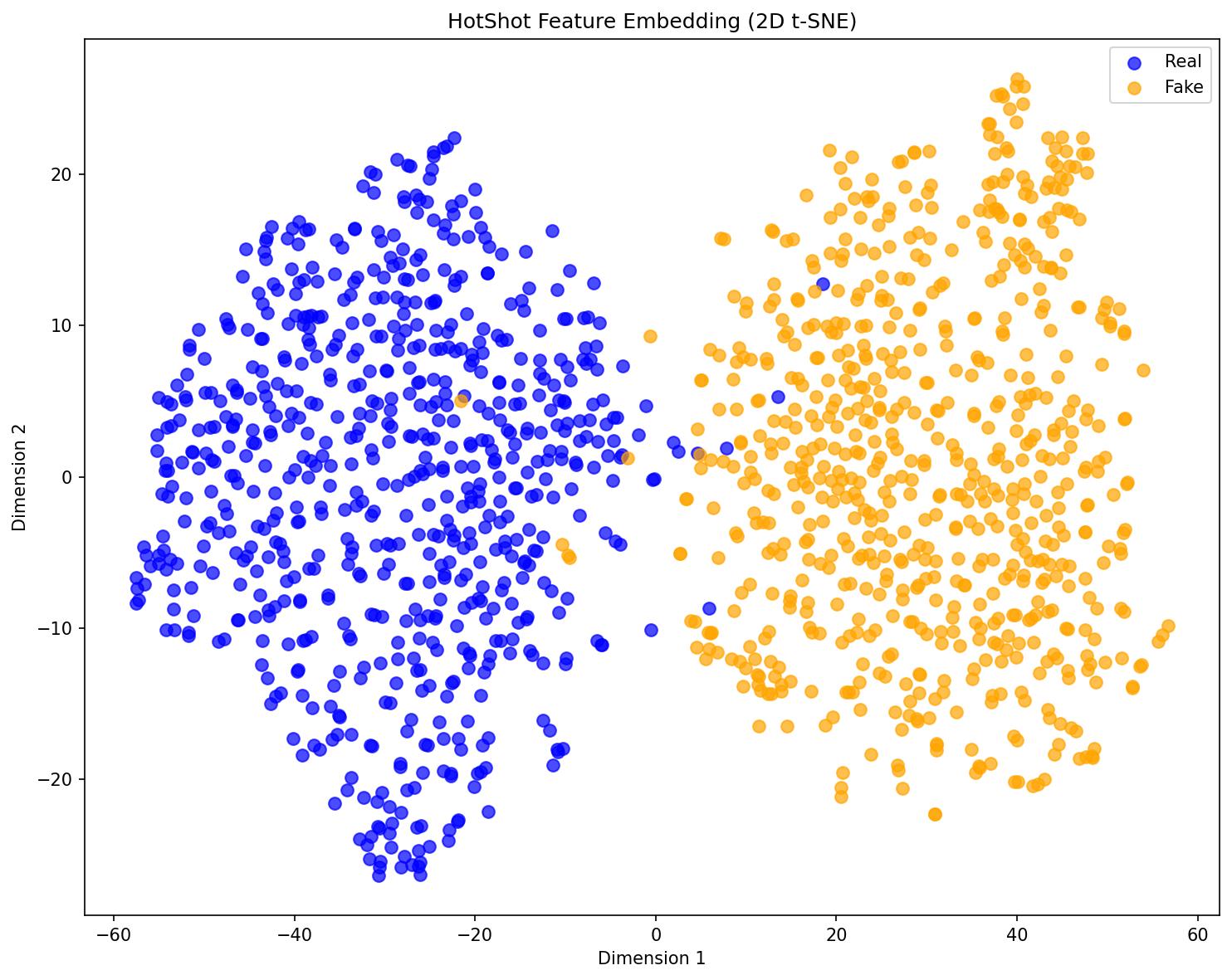}}\hfill
    \subfloat[LaVie]{\includegraphics[width=0.19\textwidth]{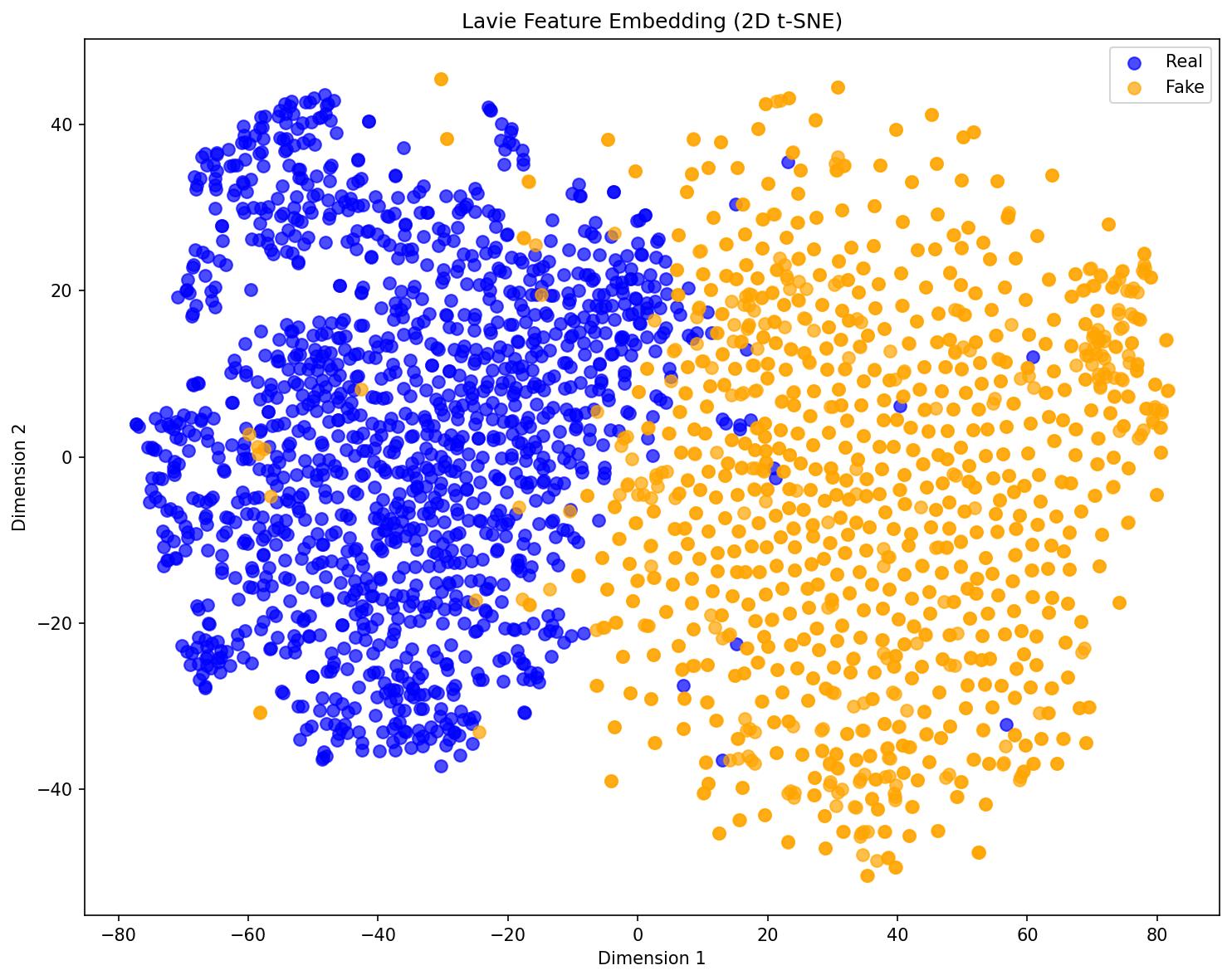}}\hfill
    \subfloat[ModelScope]{\includegraphics[width=0.19\textwidth]{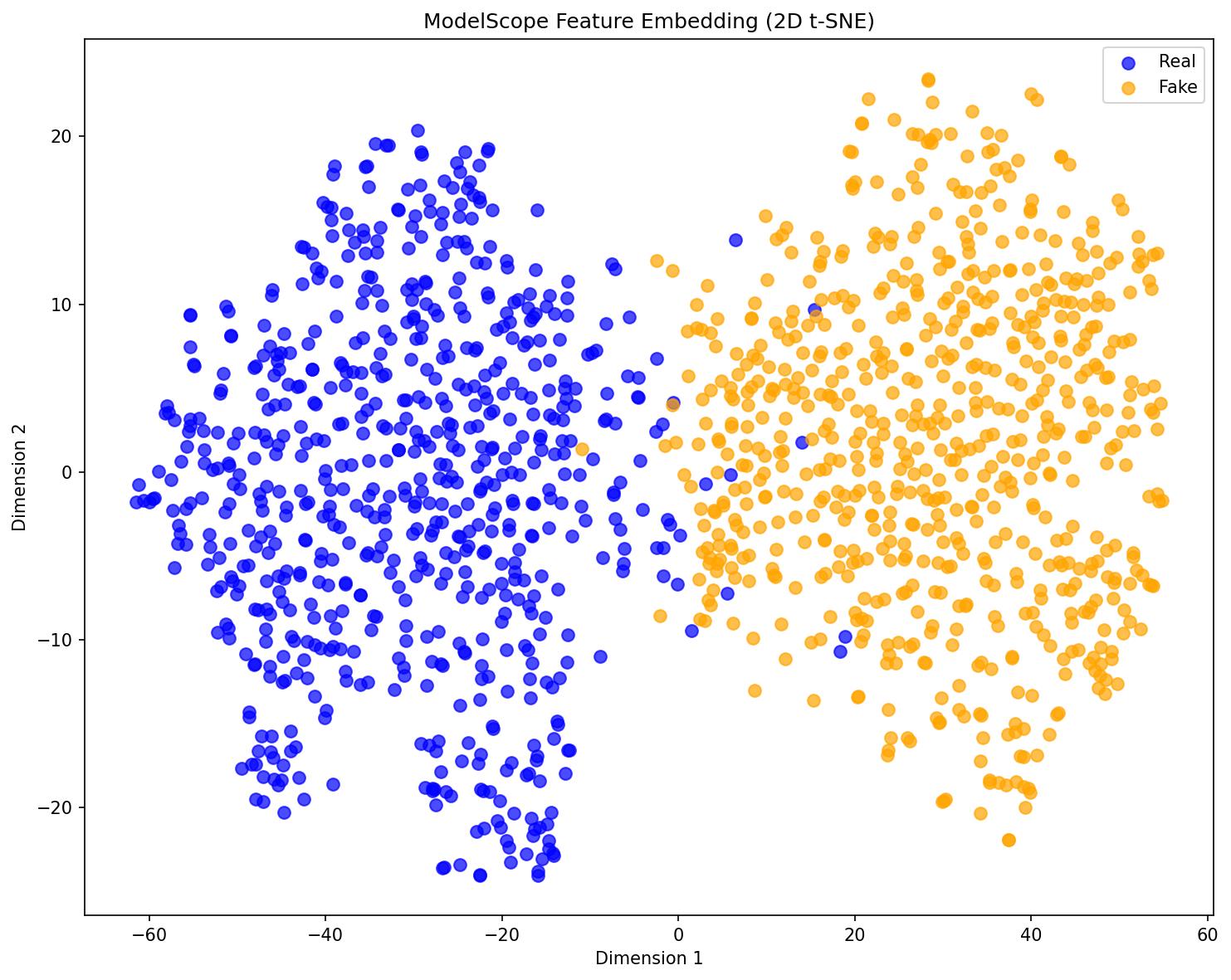}}\\[-0.2em]
    \subfloat[MoonValley]{\includegraphics[width=0.19\textwidth]{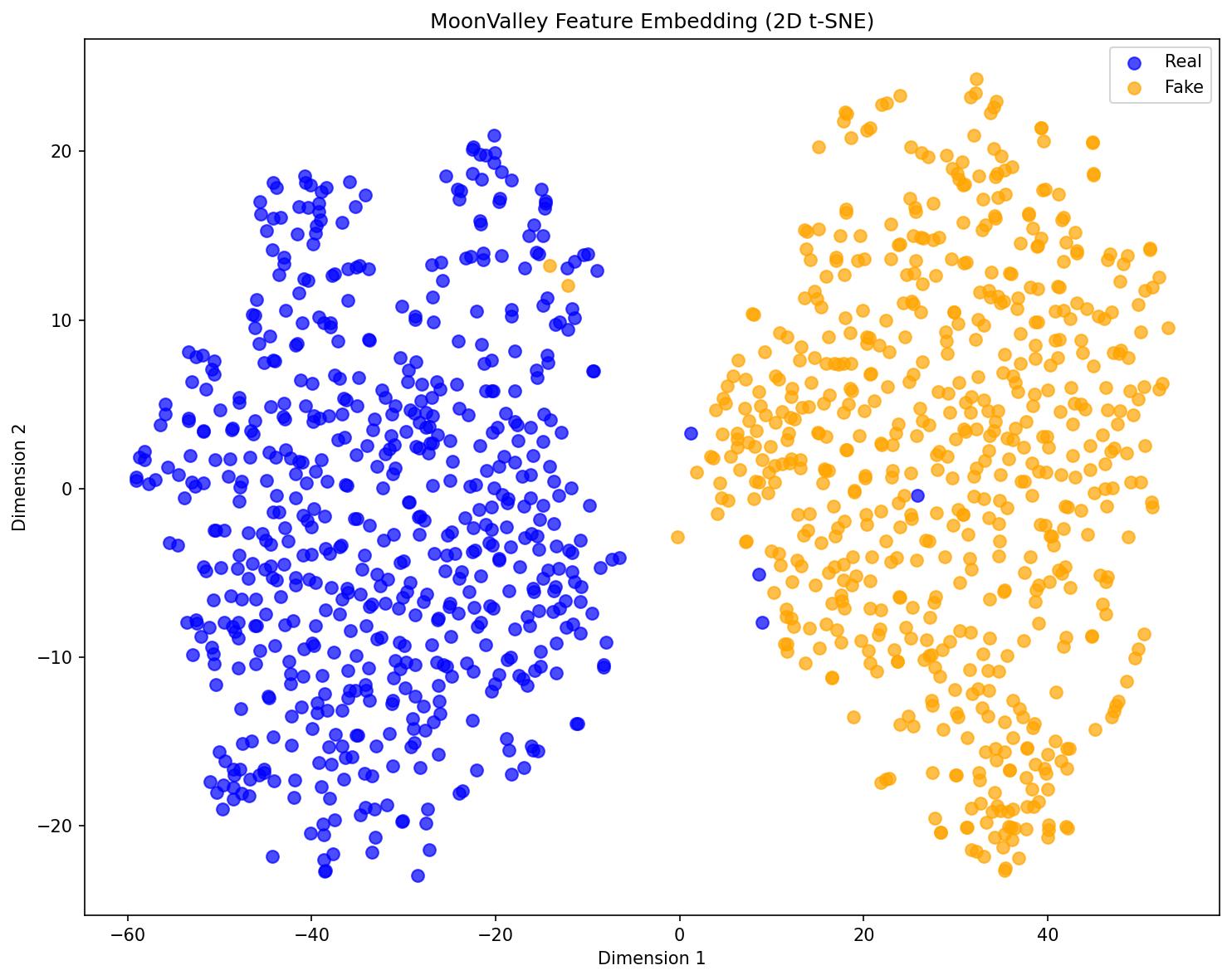}}\hfill
    \subfloat[MorphStudio]{\includegraphics[width=0.19\textwidth]{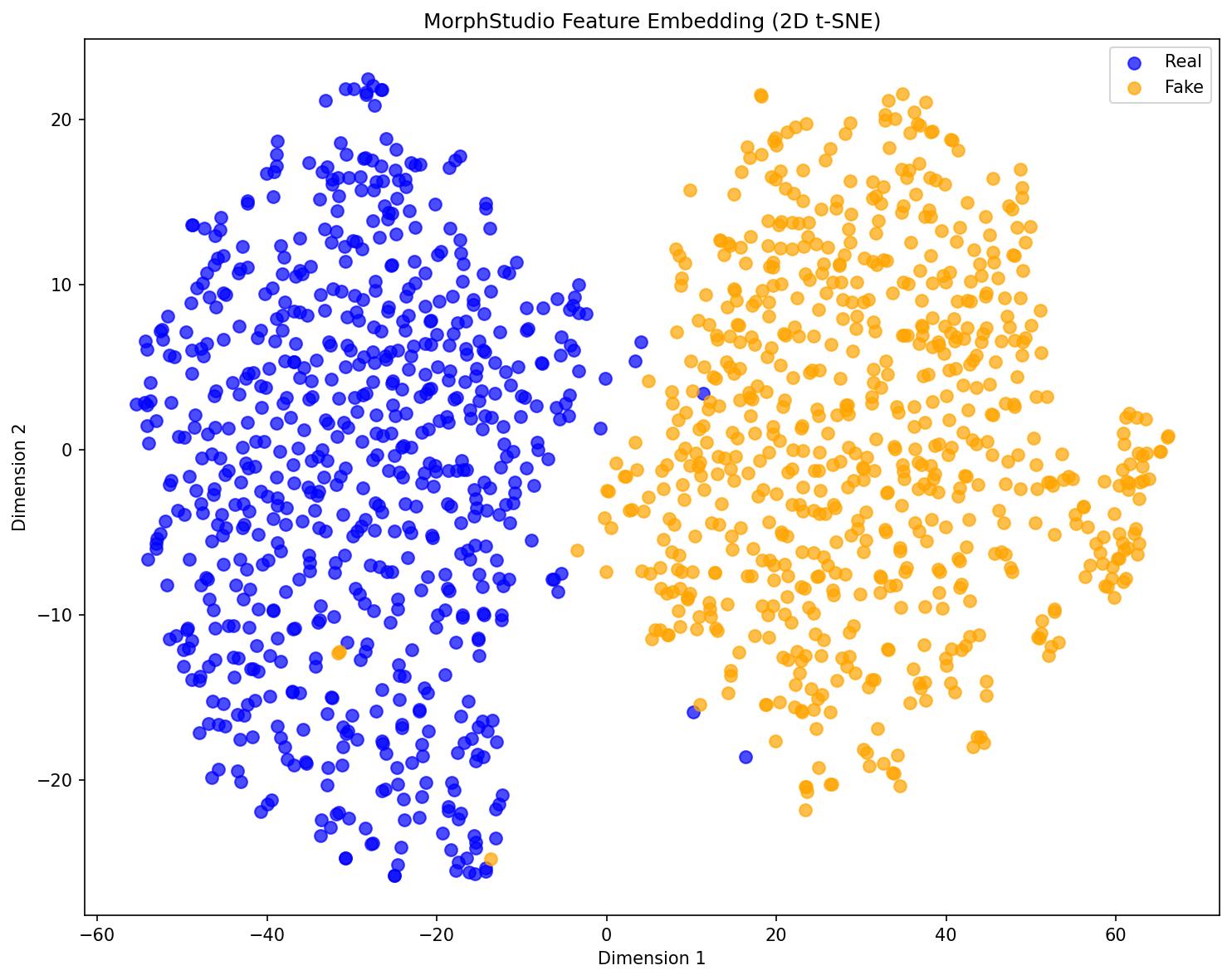}}\hfill
    \subfloat[Show-1]{\includegraphics[width=0.19\textwidth]{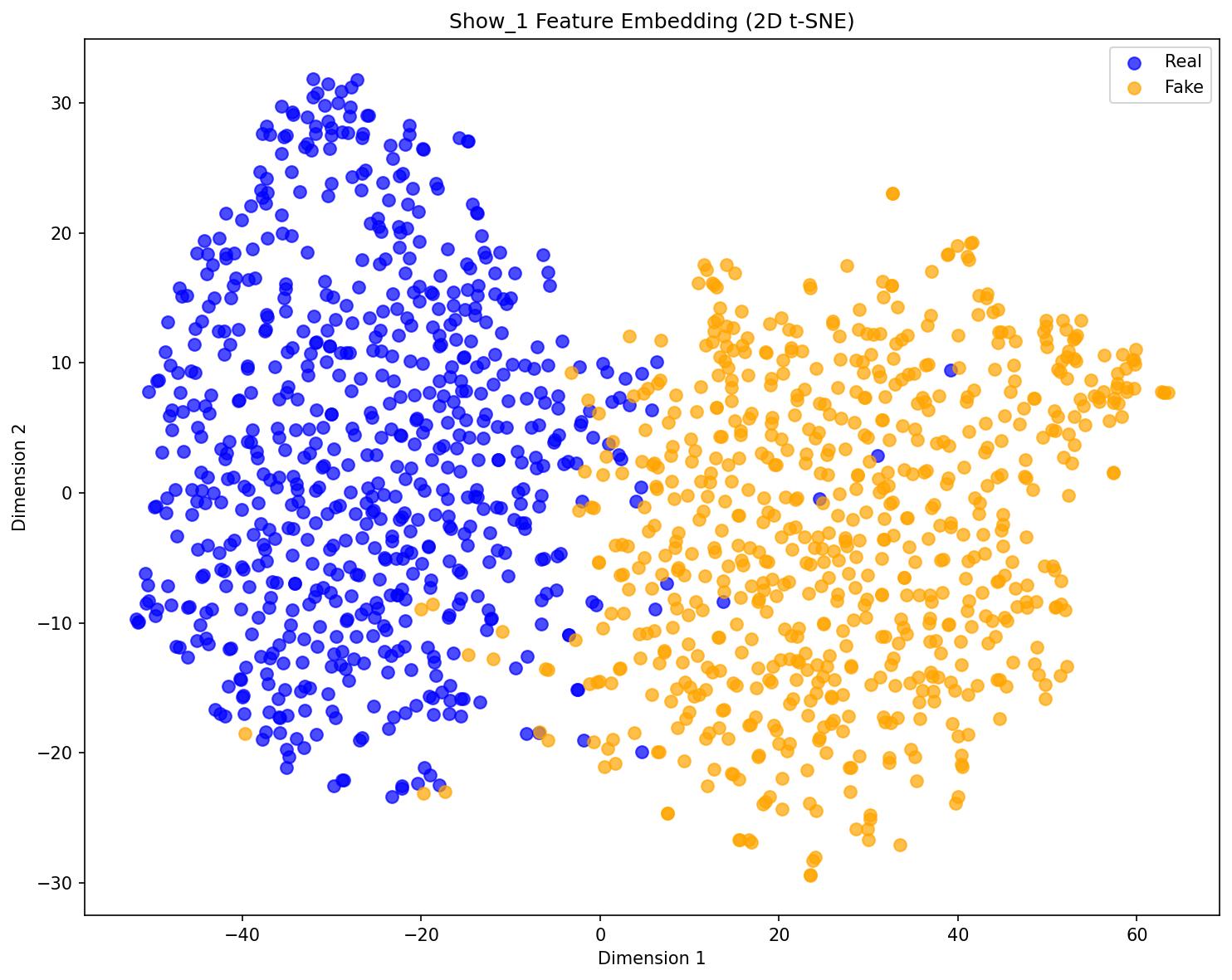}}\hfill
    \subfloat[Sora]{\includegraphics[width=0.19\textwidth]{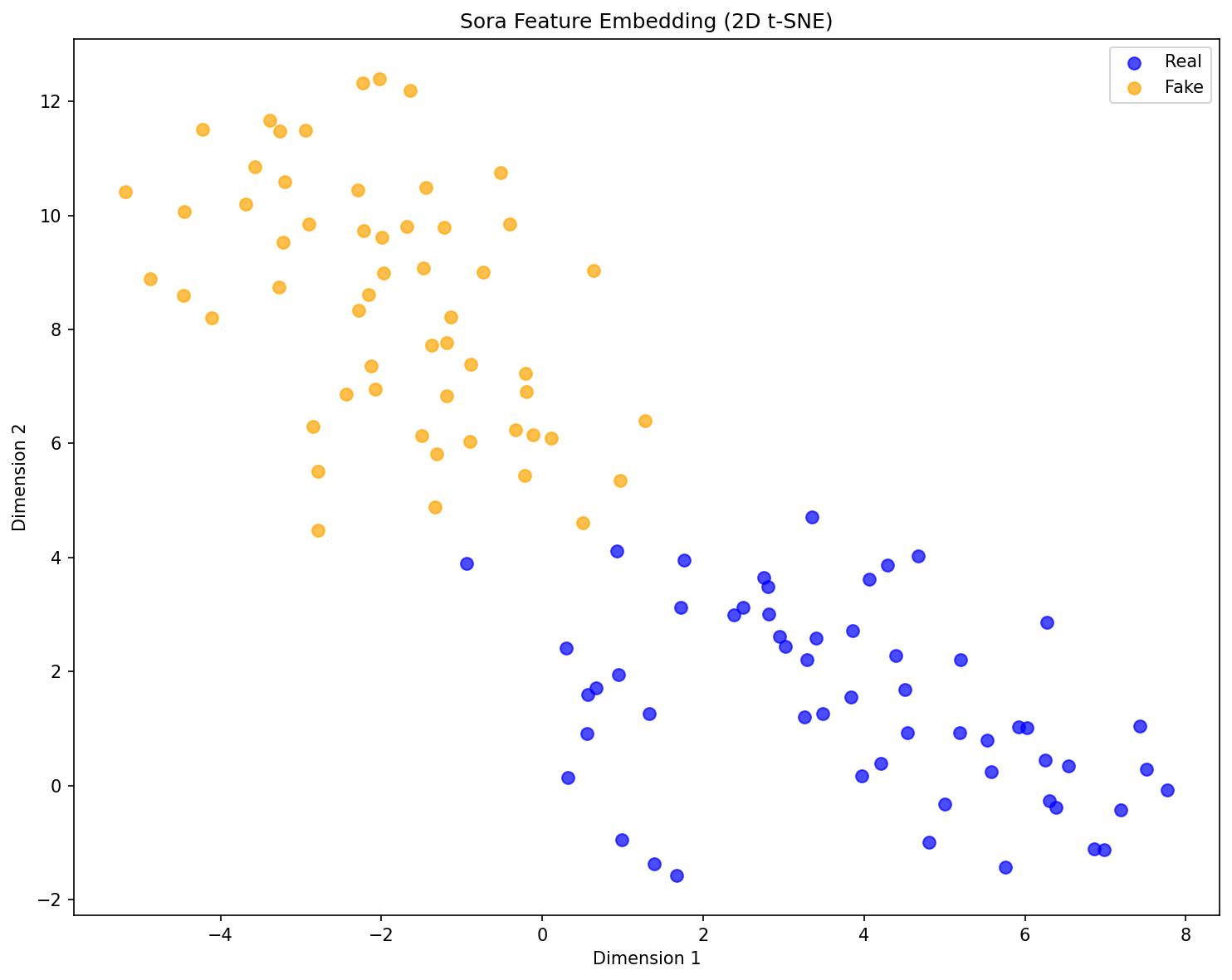}}\hfill
    \subfloat[WildScrape]{\includegraphics[width=0.19\textwidth]{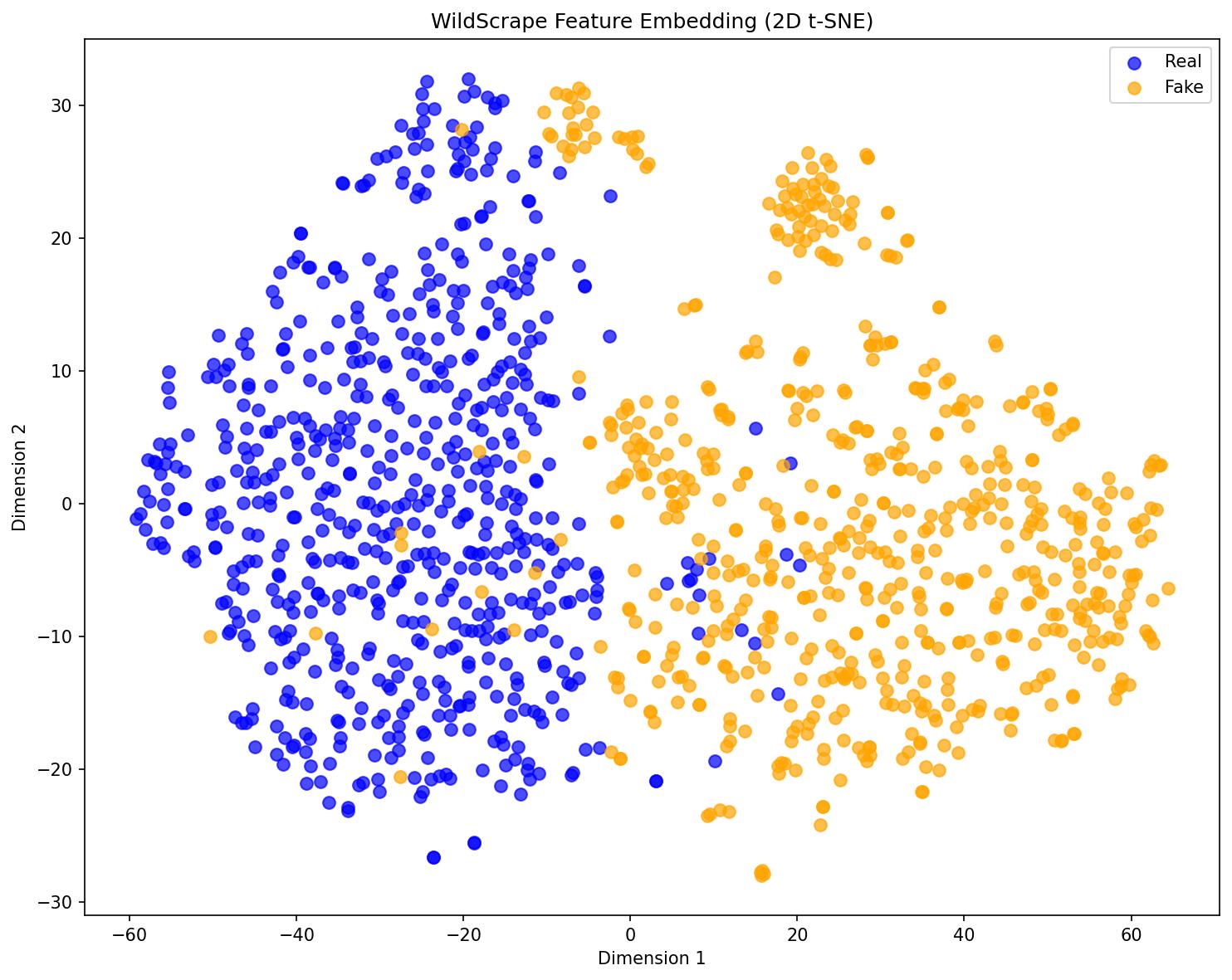}}
    \caption{t-SNE visualizations of 10 generator subsets of the GenVideo test set: (a) Crafter, (b) Gen2, (c) HotShot, (d) LaVie, (e) ModelScope, (f) MoonValley, (g) MorphStudio, (h) Show-1, (i) Sora, and (j) WildScrape. Blue and orange points denote real and AI-generated videos, respectively.}
    \label{fig:tsne_visualization}
\end{figure*}

\subsubsection{Effects of Captioning Models and Visual-Textual Encoders}
To investigate the effects of caption generation and visual-textual representation on detection performance, we evaluate different combinations of captioning models and visual-textual encoders on GenVideo.
The image-level captioners include ``BLIP-base'' and ``BLIP-large'', which generate descriptions for individual frames and are paired with CLIP-P16, CLIP-P32, XCLIP-P16, and XCLIP-P32.
The video-level captioners include ``CogVLM2-Caption'', ``UGC-VideoCaptioner'', and ``ShareGPT4Video-8B'', which generate video-level descriptions and are paired with ``LanguageBind'' or ``InternVideo~2.5''.
Tables~\ref{tab:ablation2_ap}--\ref{tab:ablation2_acc} report the corresponding AP, AUC, and ACC results.

Overall, configurations using video-level captioners and video-language encoders achieve higher mean performance than those using frame-level BLIP captioners with CLIP/XCLIP encoders.
For each video-level captioner, ``InternVideo~2.5'' consistently outperforms ``LanguageBind'' across all three metrics.
The improvements range from 1.45 to 4.04 percentage points in AP, 1.52 to 3.69 percentage points in AUC, and 2.86 to 7.16 percentage points in ACC.
These results indicate that the choice of visual-textual encoder has a substantial effect on detection performance.
In addition, ``BLIP-large'' does not consistently outperform ``BLIP-base'' under the same encoder, suggesting that increasing the captioning model size alone does not necessarily improve detection performance.

The relative performance of the video-level captioners also depends on the visual-textual encoder.
When paired with ``LanguageBind'', ``ShareGPT4Video-8B'' achieves lower mean AP and AUC than ``CogVLM2-Caption'' and ``UGC-VideoCaptioner''.
However, when paired with ``InternVideo~2.5'', it achieves the highest mean AP, AUC, and ACC of 98.74\%, 98.64\%, and 93.36\%, respectively.
The second-best configuration, ``CogVLM2-Caption'' with ``InternVideo~2.5'', achieves 98.72\% AP, 98.56\% AUC, and 93.30\% ACC, trailing the best configuration by only 0.02, 0.08, and 0.06 percentage points, respectively.
Based on its best overall performance across all three metrics, we adopt ``ShareGPT4Video-8B'' with ``InternVideo~2.5'' in the final MTOR configuration.

\begin{figure*}[ht]
\centering
\includegraphics[width=\textwidth]{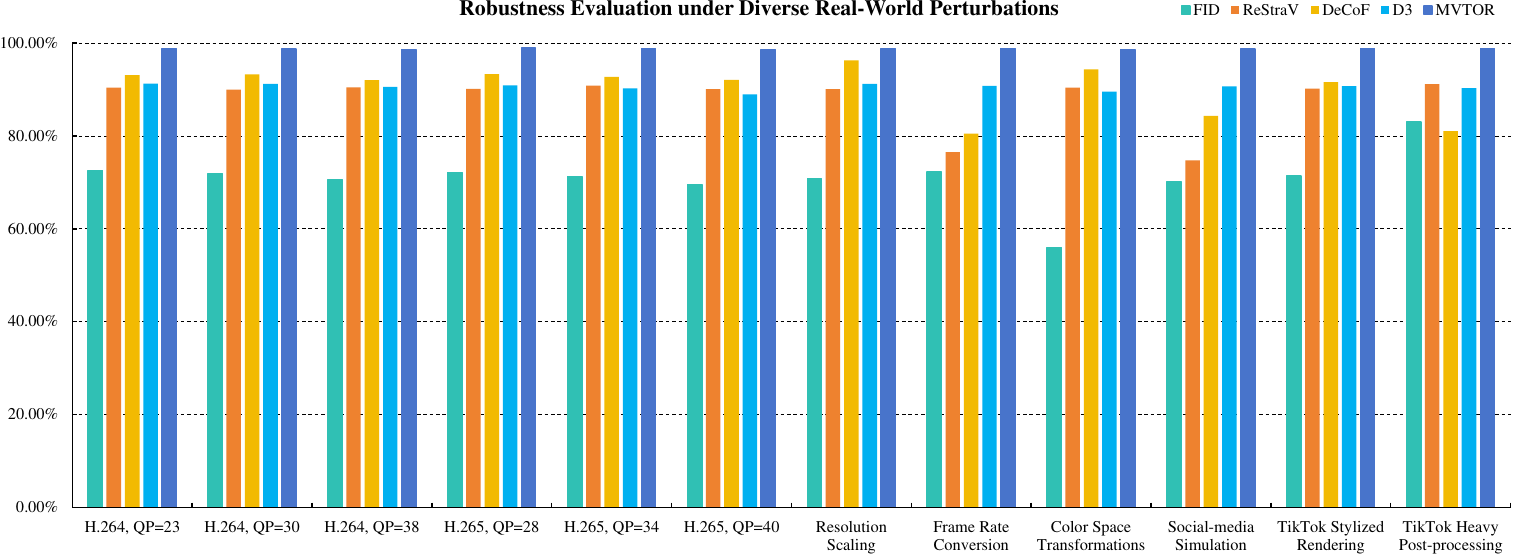} 
\caption{Robustness comparison of MTOR and four representative baselines across 12 video perturbations using AP (\%).}
\label{fig4_MTOR}
\end{figure*}

\subsection{Visualization and Robustness Analysis}
In this section, we further analyze the discriminative characteristics and practical robustness of MTOR.
Specifically, we employ t-SNE to visualize the distribution of real and AI-generated videos across the 10 generator subsets of GenVideo, providing an intuitive comparison of their separability in the representation space.
We then compare MTOR with four representative baselines under 12 practical video perturbations, covering codec compression, resolution and frame-rate changes, color-space transformation, social-media simulation, and TikTok-style post-processing.

\subsubsection{t-SNE Visualization}
To visualize the distribution of real and AI-generated videos based on the multimodal representation and the temporal descriptors used by MTOR, we perform t-SNE analysis on the 10 generator subsets of GenVideo.
Let $\mathbf{r}^{G}$ denote the last hidden-layer representation of the multimodal branch $G$.
For each video, we concatenate $\mathbf{r}^{G}$ with the standardized TOR descriptors $\tilde{\mathbf{d}}$ as $\mathbf{r}=[\mathbf{r}^{G}\|\tilde{\mathbf{d}}]$, and project the resulting representation into a 2D space using t-SNE.
This concatenation is used only for visualization, while the final MTOR prediction is obtained through score-level fusion.

As shown in Fig.~\ref{fig:tsne_visualization}, real and AI-generated samples generally form distinguishable distributions across the GenVideo generator subsets, although the degree of overlap varies among generators.
Clear class separation can be observed on Gen2, Crafter, HotShot, ModelScope, MoonValley, MorphStudio, and Sora, while LaVie, Show-1, and WildScrape exhibit relatively more overlap near the class boundaries.
Nevertheless, real and AI-generated samples remain largely distinguishable across all subsets.
The separation patterns are consistent with the high AP and AUC obtained on these subsets in Tables~\ref{tab:ap_GenVideo} and~\ref{tab:auc_GenVideo}.
In general, these visualizations provide an intuitive view of the discriminative structure jointly reflected by the multimodal representation and TOR descriptors.

\subsubsection{Robustness to Diverse Video Perturbations}
To evaluate the practical robustness of MTOR under common video transformations encountered during transmission, platform processing, and post-production, we compare MTOR with FID, ReStraV, DeCoF, and D3 under 12 perturbation settings on GenVideo.
As shown in Fig.~\ref{fig4_MTOR}, the evaluation includes H.264 compression with QP values of 23, 30, and 38; H.265 compression with QP values of 28, 34, and 40; resolution scaling; frame-rate conversion; color-space transformation; social-media simulation; TikTok stylized rendering; and TikTok heavy post-processing.
The evaluated perturbations cover both individual video transformations and compound processing pipelines, thereby providing a diverse assessment of detection robustness under practical video degradation.

MTOR consistently achieves the highest AP under all 12 perturbation settings, with an average AP of 98.79\%.
Its performance ranges from 98.60\% to 98.94\%, corresponding to a variation of only 0.34 percentage points across all settings.
In comparison, the strongest baseline D3 achieves an average AP of 90.54\%, 8.25 percentage points lower than MTOR.
The average AP values of DeCoF, ReStraV, and FID are 90.39\%, 87.91\%, and 70.98\%, respectively.

MTOR also exhibits substantially smaller performance variation across perturbations.
The AP ranges of D3, DeCoF, ReStraV, and FID are 2.30, 15.77, 16.38, and 26.94 percentage points, respectively, compared with only 0.34 percentage points for MTOR.
For example, DeCoF achieves 96.28\% AP under resolution scaling but decreases to 80.51\% under frame-rate conversion and 81.01\% under TikTok heavy post-processing.
FID is particularly sensitive to color-space transformation, where its AP decreases to 56.04\%.
By contrast, MTOR remains above 98.6\% AP under every evaluated perturbation.
These results demonstrate that MTOR maintains consistently high detection performance across diverse real-world video processing conditions.

\section{Conclusion}
In this paper, we proposed MTOR for generalizable AI-generated video detection by jointly considering video-level multimodal semantics and temporal over-regularity.
The multimodal branch represented each video from both visual and caption-derived semantic perspectives, while retaining the video itself as the sole input.
Our temporal analysis further revealed that AI-generated videos tended to exhibit stronger temporal persistence and reduced temporal variability, which we termed temporal over-regularity.
To quantify this temporal pattern, we developed the TOR component with six descriptors characterizing coarse inter-frame continuity, fine-grained token correspondence, and frame-to-video stability.
The predictions from the two branches were combined through source-calibrated score fusion using only source-domain statistics.
Across five large-scale benchmarks and 46 generator variants, comparisons with 16 representative baselines demonstrated the state-of-the-art overall performance of MTOR.
Additional ablation studies confirmed the complementary contributions of the multimodal branch and TOR component, while robustness experiments showed consistently strong performance across twelve diverse video perturbations.
Overall, these results established MTOR as an effective and robust framework for cross-generator AI-generated video detection.

\section*{Acknowledgments}
This work was supported in part by the National Natural Science Foundation of China under Grants 62441238 and U24B20185.
We acknowledge the use of the Gemini 3~\cite{google2025gemini3} for language refinement and readability enhancement of this manuscript.
It is important to clarify that this AI tool was employed solely for linguistic polishing and was not involved in conceptualization, methodology development, experimental analysis, or the derivation of scientific findings.
The authors assume full accountability for the final manuscript and affirm that all revised content complies with ethical standards, with no instances of plagiarism or research misconduct.

\bibliographystyle{IEEEtran}
\bibliography{ref}

\begin{IEEEbiography}[{\includegraphics[width=1in,height=1.25in,clip,keepaspectratio]{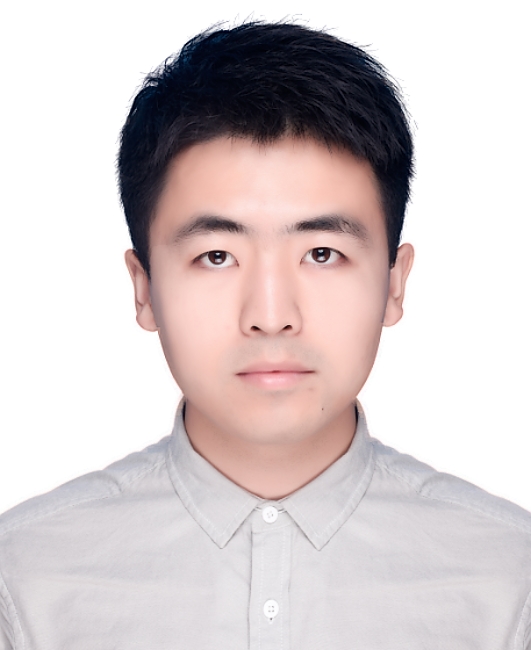}}]{Hang Wang}
received his bachelor's degree in the School of Computer Science and Technology from Xi'an Jiaotong University in 2017, and later earned his Ph.D. from the School of Cyber Science and Engineering at the same university in 2023. He also gained valuable research internship experience at Alibaba DAMO Academy from 2022 to 2023.
Currently, he is an assistant professor at the School of Automation Science and Engineering, Xi'an Jiaotong University. He is also pursuing the second Ph.D. degree with the Department of Computing at The Hong Kong Polytechnic University. His research interests include machine learning, computer vision and image processing.
\end{IEEEbiography}
\vspace{-2mm}

\begin{IEEEbiography}[{\includegraphics[width=1in,height=1.25in,clip,keepaspectratio]{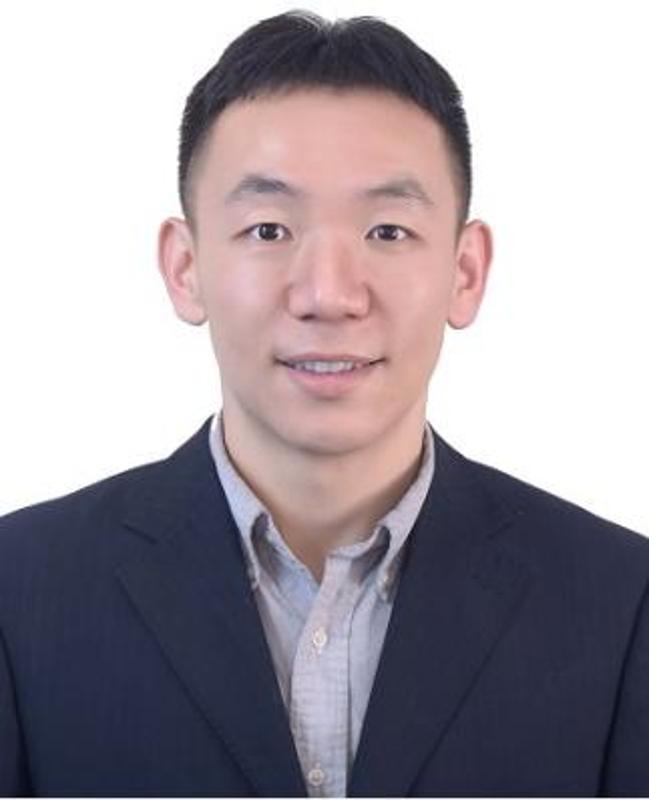}}]{Chao Shen}
(Fellow, IEEE) received the B.S. degree in automation from Xi’an Jiaotong University, China, in 2007, and the Ph.D. degree in control theory and control engineering from Xi’an Jiaotong University, China, in 2014. He is currently a Chair Professor with the School of Cyber Science and Engineering, Xi’an Jiaotong University. His research interests include AI security, insider/intrusion detection, behavioral biometrics, and measurement/experimental methodology.
\end{IEEEbiography}
\vspace{-2mm}

\begin{IEEEbiography}[{\includegraphics[width=1in,height=1.25in,clip,keepaspectratio]{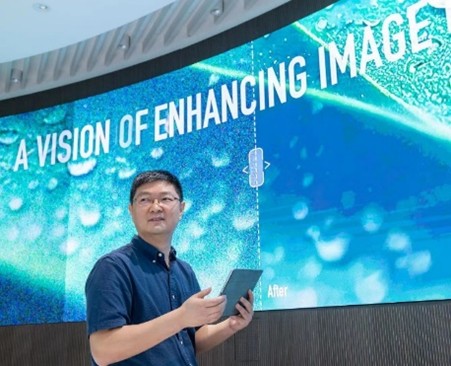}}]{Lei Zhang}
(M’04, SM’14, F’18) joined the Department of Computing, The Hong Kong Polytechnic University, as an Assistant Professor in 2006. Since July 2017, he has been a Chair Professor in the same department. His research interests include Computer Vision, Pattern Recognition, Image and Video Analysis, Deep Learning, etc. Prof. Zhang has published more than 400 papers in those areas. As of 2026, his publications have been cited more than 130,000 times in literature. Prof. Zhang is/was a Senior Associate Editor of IEEE Trans. on Image Processing, and an Associate Editor of IEEE Trans. on Pattern Analysis and Machine Intelligence, SIAM Journal of Imaging Sciences, IEEE Trans. on CSVT, and Image and Vision Computing, etc. He also served as the Senior AC, Lead AC or AC of top tier conferences such as CVPR, ICCV, ECCV, NeurIPS, ICML and ICLR. He was listed as a “Clarivate Analytics Highly Cited Researcher” consecutively from 2015 to 2025. He is an IEEE Fellow. More information can be found in his homepage \url{http://www4.comp.polyu.edu.hk/~cslzhang/}.
\end{IEEEbiography}
\vspace{-2mm}

\begin{IEEEbiography}[{\includegraphics[width=1in,height=1.25in,clip,keepaspectratio]{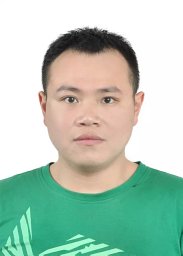}}]{Zhi-Qi Cheng} 
received his B.S. in Computer Science from Southwest Jiaotong University in 2014 and completed his Ph.D. in 2019. During his doctoral studies, he was a joint Ph.D. student at the City University of Hong Kong (2016-2017) and later at Carnegie Mellon University (2017-2019). He also gained valuable experience through internships at Alibaba DAMO Academy (2016), Google Brain (2018), and Microsoft Research (2019). From 2019 to 2022, he served as a postdoctoral research associate at the School of Computer Science of Carnegie Mellon University (CMU). He is currently a Project Scientist at the Language Technologies Institute (LTI), part of the School of Computer Science at CMU. He significantly contributed to several key projects, including DARPA's AIDA, KAIROS, IARPA's DIVA, and NIST's PSIAP. His research has had a substantial impact, such as being utilized in the Washington Post's coverage of the Capitol riots, for which he was awarded the Pulitzer Prize for Public Service. Additionally, he has been honored with the Intel Ph.D. Fellowship and IBM Outstanding Student Scholarship.
\end{IEEEbiography}

\end{document}